\pdfoutput=1

\documentclass[11pt]{article}

\usepackage[preprint]{acl}

\usepackage{times}
\usepackage{latexsym}

\usepackage[T1]{fontenc}

\usepackage[utf8]{inputenc}

\usepackage{microtype}

\usepackage{inconsolata}
\usepackage{enumitem}
\usepackage{amsmath}
\usepackage{amsfonts}
\usepackage{multirow}
\usepackage{multicol}
\usepackage{graphicx}
\usepackage{colortbl}
\usepackage{xcolor}
\usepackage{booktabs}
\usepackage{graphicx}
\usepackage{subcaption}
\usepackage{svg}
\usepackage{xspace}
\usepackage{tcolorbox}
\usepackage{tabularx}
\usepackage{adjustbox}
\usepackage{setspace}  %
\usepackage{algorithmicx}

\newcommand{\methodabbr}{Spada}           
\newcommand{\method}{\textsc{\methodabbr}\xspace}  

\title{Doubling Your Data in Minutes: Ultra-fast Tabular Data Generation via LLM-Induced Dependency Graphs}

\author{Shuo Yang \; Zheyu Zhang  \; Bardh Prenkaj \;  
         Gjergji Kasneci \vspace{0.3cm}\\
    Technical University of Munich \\
{\small \tt \{name.surname\}@tum.de}}

\begin{document}
\maketitle
\begin{abstract}
Tabular data is critical across diverse domains, yet high-quality datasets remain scarce due to privacy concerns and the cost of collection. Contemporary approaches adopt large language models (LLMs) for tabular augmentation, but exhibit two major limitations: (1) dense dependency modeling among tabular features that can introduce bias, and (2) high computational overhead in sampling.
To address these issues, we propose \method (for \underline{SPA}rse \underline{D}ependency-driven \underline{A}ugmentation), a lightweight generative framework that explicitly captures sparse dependencies via an LLM-induced graph. We treat each feature as a node and synthesize values by traversing the graph, conditioning each feature solely on its parent nodes. We explore two synthesis strategies: a non-parametric method using Gaussian kernel density estimation, and a conditional normalizing flow model that learns invertible mappings for conditional density estimation.
Experiments on four datasets show that \method reduces constraint violations by 4\% compared to diffusion-based methods and accelerates generation by nearly 9,500× over LLM-based baselines.\footnote{Our code is available at \url{https://github.com/ShuoYangtum/SPADA}}

\end{abstract}

\section{Introduction}
With the rapid advancement of data science, tabular data has become a fundamental format for storing information across diverse domains, including finance~\citep{RePEc:spr:annopr:v:339:y:2024:i:1:d:10.1007_s10479-022-04857-3}, medicine~\citep{ulmer2020trust}, cybersecurity~\citep{7307098}, and many more. Systems powered by tabular data, e.g., decision support tools~\citep{BORISOV2021100116} and anomaly detection algorithms~\citep{9810850}, have demonstrated irreplaceable value in real-world applications. Meanwhile, the high cost of data collection, coupled with privacy concerns, has rendered high-quality tabular datasets extremely scarce in practice~\citep{yang-etal-2024-p}. This scarcity underscores the urgent need for low-cost tabular augmentation methods capable of generating realistic and privacy-preserving tables.

  \begin{figure}[t]
        \centering     \includegraphics[width=.9\linewidth]{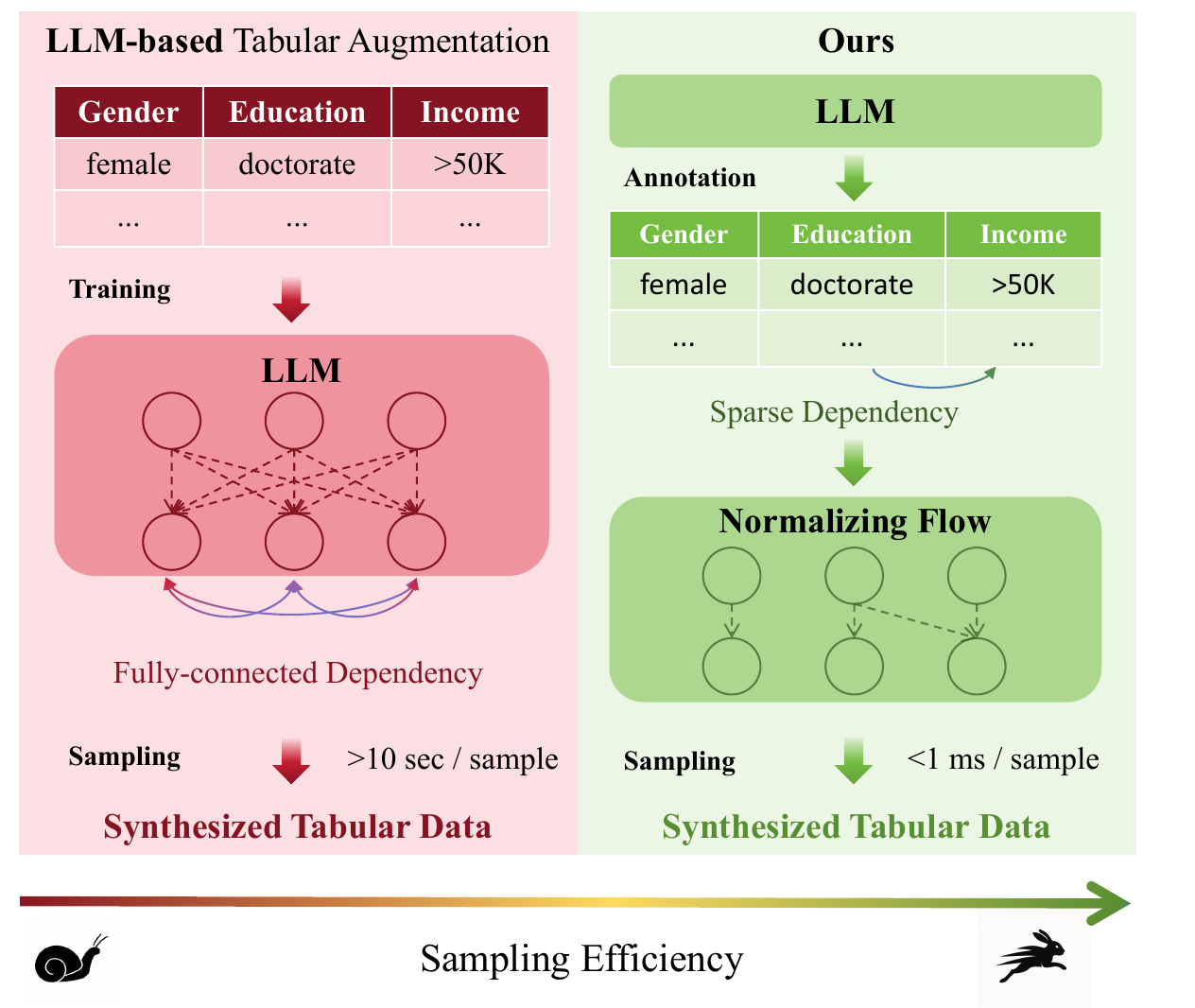}
            \caption{A comparison between \method and LLM-based approaches. Our approach leverages LLMs to annotate sparse dependency structures, effectively mitigating the bias introduced by fully connected feature assumptions in traditional methods.}
            \label{fig:compare}
            \vspace{-15pt}
    \end{figure}
    
However, the complex tabular feature dependencies pose significant challenges to high-quality generation~\citep{ren2025deeplearningtabulardata}. For example, in a population census dataset, it is rather implausible for an individual aged ``18'' to have the occupation ``professor.'' Methods that rely solely on learning statistical information from data, such as TVAE~\citep{10.5555/3454287.3454946} and CopulaGAN~\citep{Kamthe2021CopulaFF}, lack external knowledge and therefore often generate samples with logical inconsistencies~\citep{yang-etal-2024-p}. This failure to capture dependencies not only reduces the fidelity of synthetic data with respect to the real data, but also introduces unpredictable biases that increase risks when used in downstream systems~\citep{10488317}.

To capture tabular dependencies, LLMs have recently been adopted as implicit knowledge bases. Methods like GReaT~\citep{borisov2023language} pioneered this direction by converting each record into a sequence of \emph{<subject, predicate, object>}-phrases, i.e., ``\textit{feature} is \textit{value}'' templates, enabling fine-tuning of LLMs. Building on this, \citet{xu2025llmsbadsynthetictable} further optimized the ordering of these phrases to potentially strengthen these dependencies.

Despite these advancements, we argue that LLM-based methods suffer from two critical limitations:

 \noindent\textbf{Overly dense dependency modeling.} LLMs theoretically produce fully-connected information fusion among input features in hidden layers, whereas real-world entities are typically structured with sparse and heterogeneous relations~\citep{liu2023goggle}. The fine-tuning leads to unintended associations between independent features, e.g., gender and education, reflecting inherent biases in the training data~\citep{more2025assessing}. Therefore, tables generated by fine-tuned LLMs may fail to achieve the realism compared to real-world data.

\noindent\textbf{High computational cost in sampling.} LLM-based approaches require autoregressive generation of each feature value, causing the model to repeatedly pass through all layers to produce a single sample. As reported by~\citet{borisov2023language}, GPT-2~\citep{radford2019language} takes 17 seconds on average to generate one sample on two GPUs. This substantial time overhead~\citep{xia2024understanding} renders LLM-based methods impractical for realistic scenarios that demand large-scale data augmentation.

To address these limitations, we propose the \method, a novel synthesizer that explicitly models feature dependencies while dramatically reducing computational cost, as shown in Figure~\ref{fig:compare}. Specifically, we treat each feature as a node and use an LLM to extract a relational structure among them to build a directed graph~\citep{ijcai2022p0336}. We then perform a topological traversal~\citep{zheng2018dags} over this graph to sequentially generate feature values. Unlike traditional approaches, \method enforces sparsity by conditioning the synthesis of a node’s value solely on its parent nodes.
After that, we introduce two synthesis strategies instead of LLM-based generation: (1) non-parametric statistics and (2) conditional normalizing flows (NF)~\citep{flow}. 

Our contributions are summarized as follows:
\vspace{-.3cm}
\begin{enumerate}
    \item We propose a novel method for tabular augmentation. Our approach leverages LLMs to capture sparse dependencies among features and performs conditional generation using either KDE or NFs. This design decouples dependency modeling from generation, combining the structural insight of LLMs with the efficiency of lightweight models. It enables logical consistency among features while avoiding the costly autoregressive generation.
    \vspace{-5pt}
    \item \method achieves an unprecedented improvement in sampling efficiency compared to existing LLM-based baselines, reducing generation time by nearly 9,500× while improving the quality of the generated data.
    \vspace{-5pt}
    \item We conduct comprehensive experiments against SoTA methods, across four datasets, spanning binary, multi-class classification and regression tasks. The evaluation includes downstream utility, distributional alignment, privacy leakage, visual comparison, and discriminator measure. \method demonstrates remarkably reliable, scalable, and high-quality performance across the board.
\end{enumerate}

\section{Related Work}
Recent advances in tabular data generation have led to a diverse range of deep generative models, including GANs, VAEs, diffusion models and LLMs. These approaches are often evaluated along four key dimensions: utility, realism, statistical fidelity, and privacy preservation \citep{borisov2022deep, stoian2025survey}. However, a critical challenge persists in balancing these requirements while maintaining scalability and transparency.

Contemporary methods focus on end-to-end generation, learning the full joint distribution of tabular data using black-box neural architectures~\citep{tabpfn}. GAN-based models such as CTGAN \citep{10.5555/3454287.3454946}, Ganblr~\citep{zhang2021ganblr}, and Ctab-gan+ improve sample realism and utility but often suffer from mode collapse and poor interpretability. More recently, diffusion-based methods like TabDDPM \citep{kotelnikov2023tabddpm}, Findiff~\citep{sattarov2023findiff}, and TabSyn \citep{zhang2024mixedtype} have shown promise in generating high-quality samples with better coverage. Yet, they entail significant computational overhead due to their iterative sampling procedures.

A parallel line of research investigates alignment by incorporating knowledge. Approaches such as DRL \citep{stoian2025beyond} encode structural constraints through Bayesian networks or auxiliary autoencoders, but these typically require manual rule specification and are limited to discrete features. LLM-based models such as GReaT \citep{borisov2023language}, Pred-LLM~\citep{nguyen2024generating}, and LLM-TabFlow~\citep{long2025llm} attempt to model dependencies through textual representations; however, they treat generation as a monolithic text completion task, leading to excessive sampling latency and entangled dependencies.

Unlike prior work, \method decouples dependency modeling from generation. We use LLMs to extract a sparse, interpretable dependency graph, which then guides sampling via either parametric or non-parametric models. This design enhances interpretability, accelerates generation, and improves logical consistency, offering a scalable and transparent framework for structure-aware synthesis.

\section{Methodology}
\subsection{Problem Formulation}
Let $T = \{ \mathbf{t}_1, \dots, \mathbf{t}_N\} $ denote a tabular dataset with $N$ samples. Each $\mathbf{t}_i \in \mathbb{R}^M$ is a record consisting of $M$ values, i.e. $\mathbf{t}_i = \{ v_{i1}, \dots, v_{iM} \}$, where  $v_{ij}$ denotes the value of feature $f_j \in \mathcal{F}$ in sample $\mathbf{t}_i.$ Following the standard taxonomy in tabular toolbox~\citep{SDV}, we partition the feature set $\mathcal{F} = \{f_1, \dots, f_M\}$ into two disjoint subsets:
\begin{itemize}
    \item $\mathcal{F}_{\text{num}}$ contains \textbf{numerical features}, and
    \item $\mathcal{F}_{\text{cat}}$: contains \textbf{categorical features} such as boolean and discrete text strings.
\end{itemize}

A key distinction lies in their value domains. For $f_j \in \mathcal{F}_{\text{num}}$, the domain is continuous and potentially unbounded. Therefore, a generated value $v_{ij}' \in \mathbb{R}$ may not appear in $T$. On the other hand, for $f_j \in \mathcal{F}_{\text{cat}}$, the domain is a finite unordered set. Hence, any synthesized value must be drawn from the observed support: $v_{ij}' \in \mathcal{V}_j = \{v_{1j}, \dots, v_{Nj} \}.$

Our objective is to synthesize a new dataset $T' = \{ \mathbf{t}_1', \dots, \mathbf{t}_{N'}' \}$ with low computational cost, where each $\mathbf{t}_i' \notin T$  follows the same feature structure in $T$. That is, for all $j \in \{1, \dots, M\}$, the value $v_{ij}' $ corresponds to the same feature $f_j \in \mathcal{F}$ in $T$.

To ensure the realism of synthetic data, we impose the following constraints:
(1) \textbf{Distributional similarity}, that is, the marginal and joint distributions of $T'$ should match those of $T$; specifically, $\mathcal{D}_{T'} \approx \mathcal{D}_T,$ where $\mathcal{D}_T$ denotes the empirical distribution of the original dataset, and (2) \textbf{Logical consistency}, i.e., for any subset of feature values $ \{v_{ij}', \dots, v_{ik}'\} $ within a synthesized sample, there should be no semantic implausibilities or contradictions (e.g, a retired 18-years-old professor). 
    
\subsection{Dependency Graph}
In traditional LLM-based methods~\citep{borisov2023language}, fully-connected modeling typically introduce training bias~\citep{liu2023goggle}. For example, given two logically independent features $f_a$ and $f_b$, the attention weight $\text{Att}(v_a, v_b)$ between their corresponding values $v_a$ and $v_b$ should ideally approach zero. However, due to the lack of explicit constraints on feature correlations, the $\text{Att}(v_a, v_b)$ will inevitably produce positive values~\citep{attention}, resulting in a spurious correlation that distorts the feature dependency structure.

To introduce the sparse constraints among features, we employ LLMs to explicitly generate the logical dependencies. Specifically, we prompt GPT-4o~\citep{gpt4o} with the following inputs: (1) $\text{P}_\text{Intro}$, a brief description of $T$; (2) $\mathcal{F}$; and (3) $\text{P}_\text{Task}$, a task-specific instruction describing the goal of dependency identification. For each target feature $f_j \in \mathcal{F}$, the model returns a subset $\hat{F}_{f_{j}} \subseteq \mathcal{F}$ representing the features on which $f_j$ depends:
\begin{equation}
\begin{aligned}
\hat{F}_{f_{j}}=\text{LLM}(\mathcal{F}, f_j, \text{P}_\text{Intro}, \text{P}_\text{Task}),\\
\text{where } f_j \notin \hat{F}_{f_{j}} \quad \text{and} \quad \hat{F}_{f_{j}} \subseteq \mathcal{F}.
\end{aligned}
\label{eq:prompt}
\end{equation}
Here, the prompt template we used are detailed in Appendix~\ref{app:prompt}. In designing prompts, we follow the standard methodology outlined in~\citep{amatriain2024promptdesignengineeringintroduction}. For example, an output may be ``\texttt{marital status} $\rightarrow$ \texttt{age}'', indicating that ``marital status'' constrains ``age'', i.e., individuals below a certain age are unlikely to be married or divorced.

To structurally represent the generated dependencies, we construct a directed dependency graph $G = (\mathcal{F}, E)$. Here, each node $f_j \in \mathcal{F}$ is a feature, and each edge $(f \rightarrow f_j) \in E$ denotes that feature $f_i$ constrains feature $f_j$. 
In practice, for $f_j$ which is not constrained by any other feature, its dependency set $\hat{F}_{f_{j}}$ can be empty. Such features typically represent inherent properties of an entity, such as ``\textit{gender}''. To ensure that all features are integrated into a unified graph, we introduce an artificial root node $f_{\text{root}} \notin F$. For any feature $f_j$ with $\hat{F}_{f_{j}} = \emptyset$, we define a dependency from $f_{\text{root}}$ to $f_j$, resulting in the following extended definition of the edge set:
\begin{equation}
E = \bigcup_{f_j \in F} 
\begin{cases}
\left\{ (f_i \rightarrow f_j) \;\middle|\; f_i \in \hat{F}_{f_{j}} \right\}, & \text{if } \hat{F}_{f_{j}} \neq \emptyset \\
\left\{ (f_{\text{root}} \rightarrow f_j) \right\}, & \text{if } \hat{F}_{f_{j}} = \emptyset
\end{cases}
\end{equation}
Note that $f_{\text{root}}$ does not hold realistic significance. We refer readers to Appendix~\ref{app:dependency} for the complete dependencies extracted from the datasets we used.

Following~\citep{xu2025llmsbadsynthetictable}, we prioritize features that impose constraints on others, and subsequently generate the values of child nodes conditioned on their parent node values to prevent logical inconsistencies. In practice, we traverse the dependency graph $G$ starting from $f_{\text{root}}$ and follow the directed edges to generate feature values in a dependency-aware manner: 
\begin{equation}
\forall f_j \in F,\quad v_j \sim p(v_j \mid \{v_k : f_k \in \hat{F}_{f_{j}}\}).
\label{eq:dependency}
\end{equation}%
Mathematically, given $G$, we define a topological ordering over the $F$, denoted as:
\begin{equation}
f^{(1)} \prec f^{(2)} \prec \cdots \prec f^{(M)},
\label{eq:walk}
\end{equation}
where $f^{(i)} \prec f^{(j)}$ implies that $f^{(i)}$ is a parent node of $f^{(j)}$ in $G$.
During inference, we synthesize feature values following this topological order and consider the constraints from their parent nodes
\begin{equation}
\begin{aligned}
v^{(i)} \sim p(v^{(i)} \mid \{v^{(k)} \mid f^{(k)} \in \hat{F}_{f^{(i)}}\}), \quad \\
\text{for } i = 1, \ldots, M
\end{aligned}
\label{eq:cons}
\end{equation}

Nevertheless, the generated constraints $\hat{F_f}$ may introduce cycles in the graph $G$, thereby preventing the derivation of a valid topological ordering. For instance, ``latitude'' and ``longitude'' could determine the ``country'', while the ``country'' might also constrain the ranges of ``latitude'' and ``longitude''. To avoid encountering cycles, we employ an Integer Linear Programming (ILP) algorithm~\citep{ilp} to break the cycles by deleting the fewest edges that participate in any cycle of $G$. Our objective function $O_\text{ILP}$ is shown in Eq.~\eqref{eq:ILP}.
\begin{equation}
O_\text{ILP}=\quad \min \sum_{(f_i \rightarrow f_j) \in E} e_{(f_i \rightarrow f_j)}, 
\label{eq:ILP}
\end{equation}
where $e_{(f_i \rightarrow f_j)} \in \{0, 1\} \quad \forall (f_i \rightarrow f_j) \in E$. After obtaining a directed acyclic graph (DAG), we proceed to synthesize feature values for each node sequentially, as defined in Eq.~\eqref{eq:cons}. To reduce the training and sampling costs, we propose two LLM-independent strategies: a non-parametric method and a Normalizing Flow (NF) approach.

\subsection{Non-parametric Synthesis}
We propose a training-free method based on KDE to estimate conditional probabilities of a value $v$ from its similar instances in $T$. Our motivation is rooted in the fact that probabilistic models are often more effective than the neural models when the training data $T$ is not sufficient~\citep{xu2021deepnetworksreallybetter, grinsztajn2022why}. Furthermore, non-parametric synthesis eliminates the need for resource-intensive training associated with LLMs.

Specifically, for a target feature $f_j$ to be synthesized, given its dependency set $\hat{F}_{f_j}$, we filter $T$ to obtain a subset $\hat{T}$ consisting of samples that match the values of all features in $\hat{F}_{f_j}$:
\begin{equation}
\hat{T} = \left\{ t_i \in T \;\middle|\; \forall f_k \in \hat{F}_{f_j},\; v_{ik} = v^*_{k} \right\},
\label{eq:filtered_subset}
\end{equation}
where $v^*_{k}$ denotes the generated value for feature $f_k \in \hat{F}_{f_j}$ during the inference process. After that, we estimate the conditional distribution of $p(v_j \mid \hat{T})$ and sample a synthesized value from it. 

\subsubsection{Fuzzy Matching}
\label{sec:fuzzy}
However, $\hat{T}$ may be empty when $\{v^*_k\}$ are rarely observed in the original dataset. To avoid this issue, we employ a range query to relax the exact match constraint, instead of requiring all $(v^*_{k}, v_{ik})$ pairs to match exactly. For each target feature $f_j$, we define the fuzzy candidate set as:
\begin{equation}
\tilde{T} = \left\{ t_i \in T \;\middle|\; \text{Dist}\left( \mathbf{v}^*_{\hat{F}_{f_j}}, \mathbf{v}_{i,\hat{F}_{f_j}} \right) \leq \epsilon \right\}
\label{eq:fuzzy_match}
\end{equation}
Here, $\mathbf{v}^*_{\hat{F}_{f_j}}$ is the vector of previously generated conditioning values, $\mathbf{v}_{i,\hat{F}_{f_j}}$ is the corresponding vector of values in sample $t_i$, and $\text{Dist}(\cdot, \cdot)$ denotes a Hamming distance~\citep{norouzi2012fast} for categorical features in $\mathcal{F}_{\text{cat}}$ and an L1-Norm for numerical features in $\mathcal{F}_{\text{num}}$, and $\epsilon$ is the tolerance threshold controlling the fuzziness of the match.
This fuzzy matching ensures that we can always obtaine a non-empty $\hat{T}$, even under sparse conditioning combinations.

\subsubsection{BallTree}
To accelate the matching process, we construct a BallTree~\citep{omohundro1989five} on $T$ in advance, reducing the time complexity of nearest neighbor search. Specifically, for each sample $t_i \in T$, we project it onto the subspace spanned by the dependency features $\hat{F}_{f_j}$, and organize these projected vectors into a BallTree structure:
\begin{equation}
\mathcal{B}_{f_j} = \text{BallTree} \left( \left\{ \mathbf{v}_{ij}^{\hat{F}_{f_j}} \mid t_i \in T \right\} \right),
\label{eq:balltree}
\end{equation}
where $\mathbf{v}_{ij}^{\hat{F}_{f_j}}$ denotes the feature vector of sample $t_i$ restricted to the dependency set $\hat{F}_{f_j}$.

The BallTrees make our synthesis method significantly more effective, especially when $T$ is large.
By pre-building a BallTree for each $f_j \in F_{\text{num}}$ based on the corresponding dependency features, we achieve fast query times while maintaining the fidelity of conditional sampling. We refer readers to Appendix~\ref{app:balltree} for the time complexity optimization introduced by the BallTree.

\subsubsection{Kernel Density Estimation}
For $f$ in $F_{\text{num}}$, we adopt KDE, aiming to model the distribution of continuous variables and can create new, smooth values. For categorical features, we simply draw from the finite set of observed categories represented as $p(v_j \mid \hat{T})$, which is merely applicable to $F_{\text{cat}}$ with a finite set of discrete values and can only generate values previously observed in $T$. To estimate the continuous probability density of a numerical feature $f_j \in F_{\text{num}}$, we apply a Gaussian Kernel to model the distribution~\citep{bishop2006pattern}, as defined below:
\begin{equation}
\hat{p}(v_j \mid \tilde{T}) = \frac{1}{|\tilde{T}| h} \sum_{t_i \in \tilde{T}} K\left( \frac{v_j - v_{ij}}{h} \right),
\label{eq:kde}
\end{equation}
where $K(\cdot)$ is the Gaussian kernel and $h$ is the bandwidth parameter.

Finally, we generate a new sample $t'$ by sequentially assigning values to nodes following the traversal order in Eq.~\eqref{eq:walk}. This non-parametric synthesis strategy eliminates the need for model training, while enabling the generation of continuous values beyond the discrete support of the original dataset. As a result, it reduces the synthesis cost and enhances the diversity of the synthesized data.

  \begin{figure}[t]
        \centering     \includegraphics[width=\linewidth]{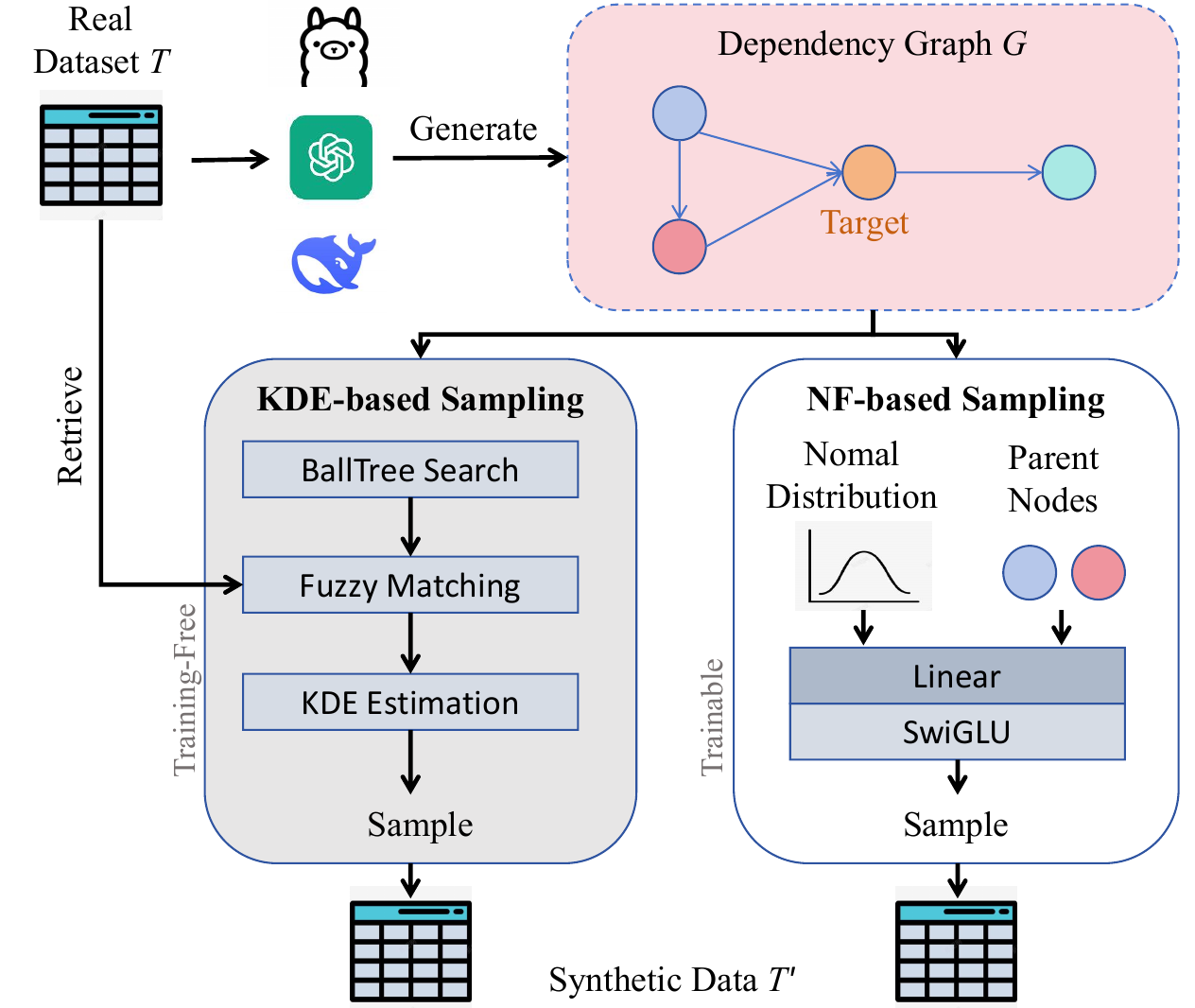}
            \caption{An overview of the proposed method. We leverage an LLM to extract dependencies among tabular features. Based on the resulting dependency graph, we obtain synthetic data using KDE or NF-based sampling.}
            \label{fig:overview}
            \vspace{-10pt}
    \end{figure}

\begin{table*}[ht]
\centering
\resizebox{.8\textwidth}{!}{%
\begin{tabular}{llrrrrrrrrrr>{\columncolor{gray!15}}r>{\columncolor{gray!15}}r>{\columncolor{gray!15}}r>{\columncolor{gray!15}}r}

\toprule
         \multirow{2}{*}{Dataset} &    & \multicolumn{2}{c}{Original}     & \multicolumn{2}{c}{TVAE}         & \multicolumn{2}{c}{CTGAN}        & \multicolumn{2}{c}{GReaT}        & \multicolumn{2}{c}{TabSyn}       & \multicolumn{2}{c}{\textbf{Ours (w/KDE)}}  & \multicolumn{2}{c}{\textbf{Ours (w/NF)}}         \\
         &    & \multicolumn{1}{c}{ACC/MSE} & \multicolumn{1}{c}{F1} & \multicolumn{1}{c}{ACC/MSE} & \multicolumn{1}{c}{F1} & \multicolumn{1}{c}{ACC/MSE} & \multicolumn{1}{c}{F1} & \multicolumn{1}{c}{ACC/MSE} & \multicolumn{1}{c}{F1} & \multicolumn{1}{c}{ACC/MSE} & \multicolumn{1}{c}{F1} & \multicolumn{1}{c}{ACC/MSE} & \multicolumn{1}{c}{F1} & \multicolumn{1}{c}{ACC/MSE} & \multicolumn{1}{c}{F1} \\ \midrule
\multirow{3}{*}{Income ($\uparrow$)}       & DT & 80.83\% & 0.73 & \textcolor{violet}{\textbf{79.50\%}}            & \underline{0.71}             & 76.27\% & 0.64 & 59.85\% & 0.60 & 77.58\% & 0.70 & 77.30\% & 0.69 & \underline{78.95\%}               & \textcolor{violet}{\textbf{0.71}}          \\
                & RF & 85.40\% & 0.78 & 82.43\% & 0.75 & 82.50\% & 0.71 & 69.42\% & 0.69 & 83.67\% & \textcolor{violet}{\textbf{0.77}}          & \textcolor{violet}{\textbf{84.18\%}}            & \underline{0.75}             & \underline{84.14\%}               & 0.75 \\
                & LR & 80.32\% & 0.67 & 78.05\% & 0.66 & 78.90\% & 0.61 & 69.57\% & \underline{0.70}             & \underline{80.21\%}               & \textcolor{violet}{\textbf{0.70}}          & \textcolor{violet}{\textbf{80.42\%}}            & 0.62 & 78.61\% & 0.56 \\ \midrule
\multirow{3}{*}{HELOC ($\uparrow$)}              & DT & 61.67\% & 0.61 & \textcolor{violet}{\textbf{64.30\%}}            & \textcolor{violet}{\textbf{0.64}}          & 63.90\% & \underline{0.64}             & 61.31\% & 0.61 & 62.13\% & 0.62 & 59.85\% & 0.60 & 63.24\% & 0.63 \\
                & RF & 71.39\% & 0.71 & 68.91\% & 0.68 & 62.78\% & 0.63 & 70.18\% & 0.70 & \textcolor{violet}{\textbf{70.73\%}}            & \underline{0.71}             & 69.41\% & 0.69 & \underline{70.58\%}               & \textcolor{violet}{\textbf{0.71}}          \\
                & LR & 69.72\% & 0.70 & 65.06\% & 0.65 & 65.47\% & 0.65 & 68.51\% & 0.68 & \underline{69.77\%}               & 0.70 & 69.57\% & \underline{0.70}             & \textcolor{violet}{\textbf{70.53\%}}            & \textcolor{violet}{\textbf{0.70}}          \\ \midrule
\multirow{3}{*}{Iris ($\uparrow$)}               & DT & 100\%   & 1.00 & 55.17\% & 0.47 & 62.07\% & 0.56 & 41.38\% & 0.36 & \underline{96.55\%}               & \underline{0.97}             & 89.66\% & 0.90 & \textcolor{violet}{\textbf{100\%}}              & \textcolor{violet}{\textbf{1.00}}          \\
                & RF & 100\%   & 1.00 & 55.17\% & 0.49 & 41.38\% & 0.37 & 44.83\% & 0.35 & \textcolor{violet}{\textbf{100\%}}              & \textcolor{violet}{\textbf{1.00}}          & \textcolor{violet}{\textbf{100\%}}              & \textcolor{violet}{\textbf{1.00}}          & \textcolor{violet}{\textbf{100\%}}              & \textcolor{violet}{\textbf{1.00}}          \\
                & LR & 100\%   & 1.00 & 62.07\% & 0.56 & 51.72\% & 0.44 & 41.38\% & 0.34 & \textcolor{violet}{\textbf{100\%}}              & \textcolor{violet}{\textbf{1.00}}          & \textcolor{violet}{\textbf{100\%}}              & \textcolor{violet}{\textbf{1.00}}          & \textcolor{violet}{\textbf{100\%}}              & \textcolor{violet}{\textbf{1.00}}          \\ \midrule
\multirow{3}{*}{Housing ($\downarrow$)} & DT & 0.14    & N/A  & 1.28    & N/A  & 5.49    & N/A  & 0.27    & N/A  & 0.29    & N/A  & \underline{0.23}                  & N/A  & \textcolor{violet}{\textbf{0.17}}               & N/A  \\
                & RF & 0.08    & N/A  & 0.52    & N/A  & 3.23    & N/A  & 0.13    & N/A  & 0.13    & N/A  & \underline{0.11}                  & N/A  & \textcolor{violet}{\textbf{0.11}}               & N/A  \\
                & LR & 0.38    & N/A  & 0.48    & N/A  & 1.62    & N/A  & \textcolor{violet}{\textbf{0.40}}               & N/A  & \underline{0.40}                  & N/A  & 0.44    & N/A  & 0.45    & N/A  \\ \bottomrule
\end{tabular}}
\caption{Performance of classifiers/regressors trained on synthetic data for downstream tasks.
\textcolor{violet}{\textbf{Bold}} indicates the best performance, and \underline{underline} indicates the second-best. ``Original'' is the original dataset $T$. ``ACC'' stands for accuracy and ``MSE'' stands for mean squared error. Besides in the original dataset, all classifiers are trained on synthetic data and tested on real ones.}
\label{tab:downstream_utility}
\end{table*}

\begin{table*}[ht]
\centering
\caption{Violation rates $\downarrow$. We present measurements with 95\% confidence interval. \textcolor{violet}{\textbf{Bold}} indicates the best performance, and \underline{underline} indicates the second-best.}
\label{tab:violation_rates}
\resizebox{.8\textwidth}{!}{%
\begin{tabular}{lcccccc}
\toprule
Dataset & \multicolumn{1}{c}{TVAE} & \multicolumn{1}{c}{CTGAN} & \multicolumn{1}{c}{GReaT} & \multicolumn{1}{c}{TabSyn} & \textbf{Ours (w/KDE)} & \textbf{Ours (w/NF)}  \\ \midrule 
Income & 4.21 $\pm 0.64$\%                  & 34.31 $\pm1.18$\%                   & \textcolor{violet}{\textbf{0.00 $\pm 0.00$\%}}                   & 2.32 $\pm 0.49$\%            & 3.56 $\pm 0.98$\%                & \textcolor{violet}{\textbf{0.00 $\pm 0.00$\%}}         \\
Housing & 15.55 $\pm 0.55$\%                  & 34.48 $\pm 0.72$\%                   & \underline{3.61 $\pm 0.28$\%}              & 10.70 $\pm 0.72$\%           & 5.26 $\pm 0.62$\%                & \textcolor{violet}{\textbf{1.37} $\pm 0.18\%$} \\ \midrule 
Mean ($\downarrow$)    & 9.88\%                  & 34.40\%                   & \underline{1.81\%}                   & 6.51\%                     & 4.41\%          & \textcolor{violet}{\textbf{0.69\%}}     \\ \bottomrule[0.8pt]
\end{tabular}
}
\vspace{-10pt}
\end{table*}

\subsection{Conditional Normalizing Flow}
\label{sec:NF}
The non-parametric synthesis method achieves a theoretically lower computational cost compared to LLM-based generation. However, when the size of $T$ is small, the number of matching samples in $\hat{T}$ may be insufficient, leading to biased probability estimates under the fuzzy matching strategy. Conversely, when $T$ is large, repeated access to the dataset during sampling increases the sampling overhead.
To address this, we introduce a parametric generative method based on conditional NFs, enabling efficient synthesis without data access.

\subsubsection{Theoretical Framework}
NFs transform a standard Gaussian distribution into the probability density of our target feature value $v$ through a sequence of differentiable mappings. Specifically, let $z \sim p_Z(z)$ be a latent variable sampled from the standard Gaussian distribution, and let $f_\theta$ be a learnable transformation parameterized by $\theta$. The target feature value $v$ is then given by:
\begin{equation}
v = f_\theta(z \mid \{v_k \mid f_k \in \hat{F}_f\}),
\end{equation}
where the transformation is conditioned on the values of features in $\hat{F}_f$. We encode categorical features in $F_{\text{cat}}$ by using a label encoder~\citep{labelencoder} into continuous representations. The conditional density is:
\begin{equation}
p(v \mid \{v_k\}) = p_Z(f_\theta^{-1}(v \mid \{v_k\})) \left| \det \left( \frac{\partial f_\theta^{-1}}{\partial v} \right) \right|,
\label{eq:log_likelihood}
\end{equation}
where the Jacobian determinant~\citep{pmlr-v37-rezende15} captures the local volume change induced by the transformation.

\subsubsection{Implementation Strategy}
In training, we maximize the likelihood of observed feature values under the modeled conditional distribution. The loss funtion is shown in Eq.~\eqref{eq:loss}:
\begin{equation}
\mathcal{L}(\theta) = - \frac{1}{N} \sum_{i=1}^N \log p(v \mid \{v_k\}; \theta).
\label{eq:loss}
\end{equation}%
In practice, we parameterize $f_\theta$ using a fully connected neural layer with SwiGLU~\citep{swiglu}. The model takes as input the latent variable $z$ and the conditioning values $\{v_k\}$, and outputs the synthesized value $v$. This NF enables learning expressive conditional distributions over both continuous and categorical features, facilitating high-fidelity and efficient data synthesis. An overview of our framework is shown in Figure~\ref{fig:overview}.

\begin{figure*}[ht]
    \centering
    \includegraphics[width=.95\textwidth]{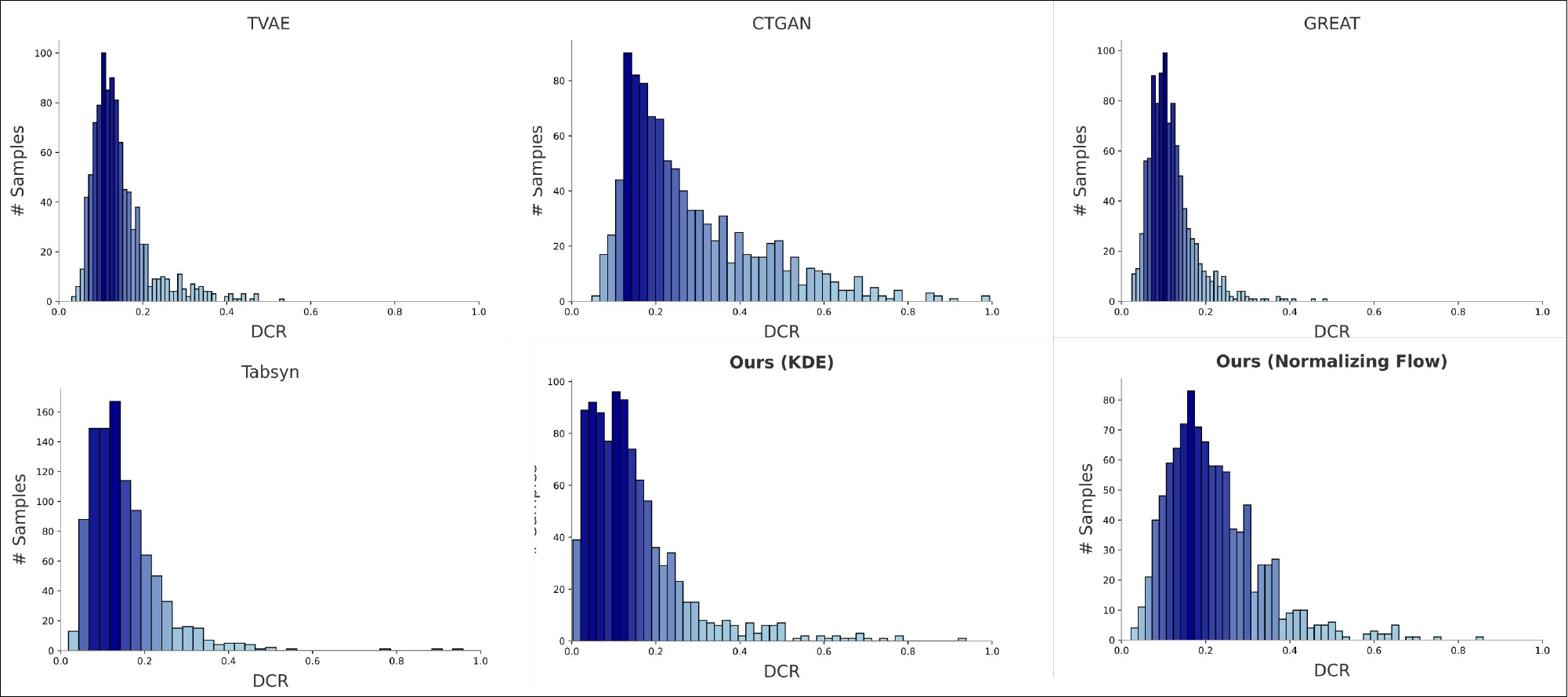}
    \caption{DCR for the California Housing dataset, evaluated with respect to the original training set. A smaller DCR suggests that the model may overfit and copy certain feature values from the original data.}
    \label{fig:privacy_protection}
\end{figure*}

\section{Experiments}
\subsection{Datasets}
\textbf{Binary Classification.} The 
\emph{Adult Income} dataset \citep{adult_2} contains 16 demographic and occupational features, which are used to predict an individual's annual income level.  
The \emph{Home Equity Line of Credit} (HELOC) dataset~\citep{heloc} includes 24 credit-related features extracted from people's credit reports. The task is to predict whether an individual will repay their HELOC amount within the next two years.

\noindent\textbf{Multi-class Classification.} The well-known
\emph{Iris} dataset~\citep{iris_53} comprises four numerical features describing the sepal and petal dimensions of iris flowers. The task is to classify the sample into an iris species.

\noindent\textbf{Regression.} The 
\emph{California Housing} dataset~\citep{house} consists of 10 features related to housing and geographic attributes. Our task is to predict the latitude and longitude of a property.

\subsection{Evaluation Metrics}
\noindent\textbf{Downstream Utility.} 
We generated synthetic datasets of the same size as the original datasets and trained decision tree (DT), random forest (RF), and logistic regression (LR) models as classifiers and regressors, following the respective tasks of the datasets used. For classification, we reported accuracy and F1, while for regression tasks, we reported Mean Squared Errors. The results are presented in Table~\ref{tab:downstream_utility}.

\noindent\textbf{Privacy Protection.} 
Following~\citep{zhang2024mixedtype}, we used the L1 norm to calculate the Distance to Closest Records (DCR) to $T$. We normalize each column and compute the average. A high DCR suggests minimal overlap between the feature values of synthetic data and those in the original dataset, which contributes to stronger privacy preservation. The results are presented in Figure~\ref{fig:privacy_protection}.

\begin{figure*}[ht]
    \centering
    \includegraphics[width=.95\textwidth]{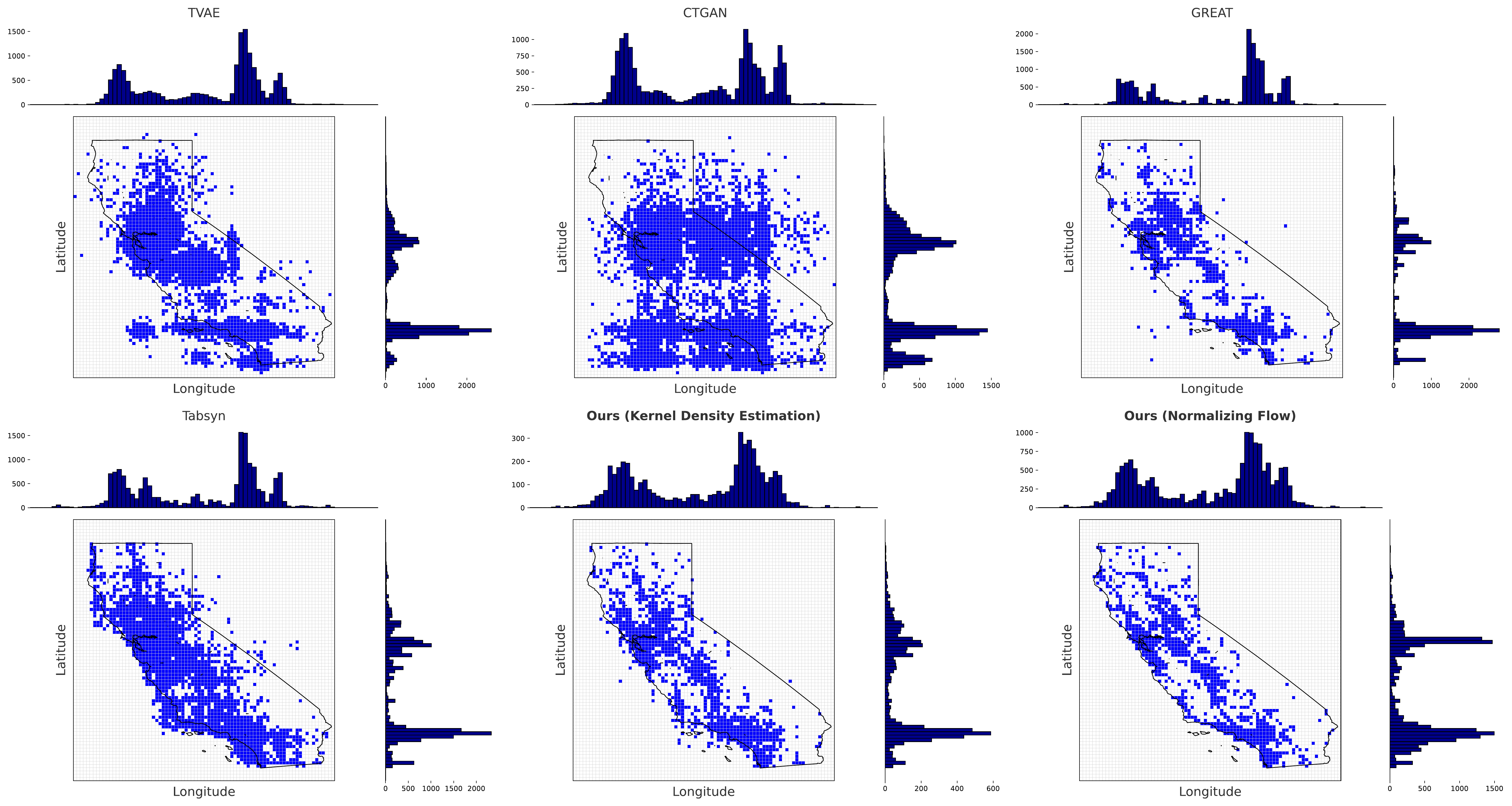}
    \caption{Comparison of the generated samples for the California Housing dataset, which includes characteristic information about various properties in California, USA. Joint histogram plots of the highly correlated variables Latitude and Longitude are shown. The black outline represents the true boundary of the state of California.}
    \label{fig:california_comparison}
\end{figure*}

\noindent\textbf{Data Fidelity.} 
 Following~\citep{xu2025llmsbadsynthetictable}, we compute the two kinds of violation rates pre-defined on the Income and California Housing datasets. For the Income dataset, the violation rate refers to the proportion of generated samples exhibiting inconsistencies between the \texttt{``educational-num''} and \texttt{``education''} features. For the Housing dataset, it refers that of falling outside the geographic boundaries of California. The results are presented in Table~\ref{tab:violation_rates}. Additionally, we visualized the geographic coordinates of the synthetic samples generated from the California Housing dataset, as shown in Figure~\ref{fig:california_comparison}.

\begin{table}[ht]
\centering
\resizebox{\linewidth}{!}{%
\begin{tabular}{llcccccc}
\toprule
Dataset                  &               & \multicolumn{1}{c}{TVAE} & \multicolumn{1}{c}{CTGAN} & \multicolumn{1}{c}{GReaT} & TabSyn & \textbf{Ours (w/KDE)}    & \textbf{Ours (w/NF)}     \\ \midrule
\multirow{2}{*}{Income}  & Training & 57 sec                   & 108 sec                   & 6h 10m 34 sec             & 34 min & \textcolor{violet}{\textbf{0 sec}}              & \underline{36 sec}          \\
                         & Sampling & \textcolor{violet}{\textbf{\textless{}1 ms}}          & \textcolor{violet}{\textbf{\textless{}1 ms}}           & 9 sec                     & 5 sec  & 16 ms           & \textcolor{violet}{\textbf{\textless{}1 ms}} \\ \midrule
\multirow{2}{*}{HELOC}   & Training & \underline{29 sec}                   & 41 sec                    & 1h 47min 38sec            & 29 min & \textcolor{violet}{\textbf{0 sec}}             & 45 sec          \\
                         & Sampling & \textcolor{violet}{\textbf{\textless{}1 ms}}          & \textcolor{violet}{\textbf{\textless{}1 ms}}           & 45 sec                    & 2 sec  & 65 ms           & 2 ms  \\ \midrule
\multirow{2}{*}{Iris}    & Training & \underline{\textless{}1s}           & 3 sec                     & 17 sec                    & 14 min & \textcolor{violet}{\textbf{0 sec}}              & \underline{\textless{}1s}   \\
                         & Sampling & \textcolor{violet}{\textbf{\textless{}1 ms}}          & \textcolor{violet}{\textbf{\textless{}1 ms}}           & 4 sec                     & 254 ms & \textcolor{violet}{\textbf{\textless{}1 ms}} & \textcolor{violet}{\textbf{\textless{}1 ms}} \\ \midrule
\multirow{2}{*}{Housing} & Training & \underline{33 sec}                   & 50 sec                    & 1h 18min 8sec             & 30 min & \textcolor{violet}{\textbf{0 sec}}              & 1min 8 sec      \\
                         & Sampling & \textcolor{violet}{\textbf{\textless{}1 ms}}          & \textcolor{violet}{\textbf{\textless{}1 ms}}           & 8 sec                     & 2 sec  & 74 ms           & 3 ms \\ \bottomrule
\end{tabular}
}
\caption{Average training time and sampling time per instance. The devices we used are shown in Appendix \ref{sec:hardware}. \textcolor{violet}{\textbf{Bold}} indicates the best performance.}
\label{tab:time_cost}
\end{table}

\section{Results}

\subsection{Computational Cost and Time Cost}
Table~\ref{tab:time_cost} presents the training time and the average sampling time per sample for both the baseline methods and our proposed approaches.

\subsection{Discussion}
\method consistently outperforms all the baselines in terms of downstream utility, as shown in Table~\ref{tab:downstream_utility}. The most striking result is observed on the multi-class Iris dataset, where all three classifiers trained by our NF-based method achieve perfect accuracies. We analyze that the small sample size and low feature dimensionality of the Iris dataset hinder the ability of LLM-based methods such as GReaT to effectively learn the underlying data distribution, thereby resulting in suboptimal performance. Moreover, we observe that the NF-based approach slightly outperforms the KDE-based one on average. This supports our hypothesis mentioned in \S\ref{sec:NF}: \textit{with small size $\hat{T}$, fuzzy matching can lead to biased density estimation, making it less accurate than trainable neural networks}.

In terms of privacy protection, we observe that the DCR values of synthetic data generated by most baselines are close to zero, as shown in Figure~\ref{fig:privacy_protection}. The most severe overlap is observed in our KDE-based method. Since KDE estimates probability densities by retrieving from the original dataset, it inevitably results in a large amount of similar feature values. Similarly, the DCR means for synthetic data produced by TVAE, GReaT, and TabSyn are all below 0.1. In contrast, our NF-based method and CTGAN achieve average DCR values around 0.2, indicating better privacy preservation compared to other methods.  \citet{mendelevitch2021fidelityprivacysyntheticmedical} reported that synthetic datasets with higher DCR values exhibit reduced risks of unintended memorization and re-identification. Therefore, increasing the DCR from 0.1 to 0.2 extremely enhances privacy protection by doubling the distinguishability between synthetic and real records, thereby reducing the likelihood of re-identification attacks. This aligns with established research indicating that higher DCR values contribute to stronger privacy safeguards. However, we note that the high DCR of CTGAN may stem from its difficulty in faithfully capturing the underlying distribution of the real data.

Figure~\ref{fig:california_comparison} illustrates that \method faithfully captures the real data distributions. Notably, CTGAN exhibits the most severe boundary violations, where the spatial outline of California becomes unrecognizable. Compared to baselines, \method produces samples that remain largely within valid geographic boundaries, indicating that they better model the relationship between latitude and longitude.
These findings are further supported by Table~\ref{tab:violation_rates}. Our NF-based method achieves a 2\% reduction in the violation rate compared to the LLM-based GReaT model on the Housing dataset. This substantial improvement highlights the effectiveness of incorporating sparsity-aware dependency structures during data generation. Unlike fully-connected generative strategies adopted by LLM-based methods, our approach explicitly models and respects the true conditional dependencies among features, thereby ensuring higher logical fidelity in synthesized data.

\method demonstrates high efficiency in both training and sampling stages, as shown in Table~\ref{tab:time_cost}. Compared to parameter-heavy models such as \textsc{GReaT} and \textsc{TabSyn}, our NF-based model reduces training time by over 100$\times$ and approximately 30$\times$, respectively. During sampling, the NF-based model synthesizes each sample in under one millisecond, achieving efficiency comparable to \textsc{TVAE} and \textsc{CTGAN}. Remarkably, our model is on average 9,500$\times$ faster than the LLM-based \textsc{GReaT}, highlighting its potential for large-scale data augmentation.
On the other hand, although our KDE-based method is slightly slower than the NF-based approach, it estimates the need of training and thus advantageous in scenarios with limited computational resources.

\section{Conclusion}
We proposed \method, a novel and lightweight framework for tabular augmentation that disentangles dependency modeling from data generation. By leveraging LLMs to extract sparse feature dependencies and employing lightweight generators, our approach significantly improves the fidelity of synthetic data by 2\% while achieving 9,500x speedup in sampling. Extensive experiments across multiple datasets and evaluation metrics demonstrate the effectiveness of \method in maintaining downstream utility, enhancing realism, and ensuring privacy preservation. Overall, our findings underscore the potential of leveraging LLM-derived structural priors in conjunction with lightweight generative models for scalable, high-fidelity, and privacy-preserving tabular synthesis. Future work may explore domain-specific adaptations and further integration with interdisciplinary evaluation frameworks to better assess the societal impact of synthetic data technologies.

\section{Limitation}
\label{sec:limitation}
While \method significantly enhances the effectiveness of LLM-based approaches and greatly reduces the cost of data augmentation, it still has two key limitations:

\paragraph{1) Reliance on the quality of LLM-annotated dependencies.}
Both of our synthesis strategies rely on the dependency graph produced by an LLM. As such, the accuracy of this graph may theoretically affect downstream model performance. Nevertheless, recent studies have demonstrated the reliability of LLMs in annotation tasks~\citep{gilardi2023chatgpt}, and in our experiments, the GPT-4o-generated dependency graphs enabled our models to achieve strong performance across multiple datasets. Therefore, although this limitation exists in theory, we have not observed significant evidence that it affects practical performance.

\paragraph{2) Inability to model complex data types.}
\method, following convention, categorizes tabular values into categorical and numerical features. However, for multimodal datasets that include images, videos, or other high-dimensional data types, our current framework is not applicable. Moreover, categorical features are assumed to be finite and text-representable. As a result, \method cannot generate open-domain text or novel tokens outside the original dataset; instead, it selects from a fixed set of observed values. We also note that this limitation is shared by existing methods such as TVAE and SynTab, which are similarly restricted to structured tabular data.

Despite these limitations, \method remains broadly applicable and offers significant improvements in efficiency and effectiveness over existing approaches. The observed limitations either have minimal empirical impact or are inherent to the general problem setting, rather than specific to our solution. Therefore, we argue that they do not diminish the core contributions of our work.

\section*{Acknowledgments}
The authors acknowledge the use of ChatGPT solely to refine the text in the final manuscript.

\bibliography{anthology,custom}

\newpage

\appendix
\section{Theory Analysis}

\subsection{Asymptotic Consistency of the KDE with Gaussian Kernel}

\paragraph{Problem Setup}
Given a target feature \( f_j \in \mathcal{F} \) and its parent features \( \hat{F}_{f_j} \subseteq \mathcal{F} \). For the dataset \( T = \{\mathbf{t}_i\}_{i=1}^n \), we first retrieve a subset \( \hat{T} \) via fuzzy matching on \( \hat{F}_{f_j} \):
 \begin{equation}
\hat{T} = \left\{ \mathbf{t}_i \in T \, \bigg| \, \text{Dist}\left(v_{i,\hat{F}_{f_j}}, v_{\hat{F}_{f_j}}^*\right) \leq \epsilon \right\},
 \end{equation}
where \( v_{\hat{F}_{f_j}}^* \) are the parent feature values generated, and \(\text{Dist}(\cdot)\) combines Hamming distance for categorical features and L1 distance for numerical features, as we mentioned in \S\ref{sec:fuzzy}. 

The conditional density of \( f_j \) is then estimated via Gaussian KDE on \( \hat{T} \):
\begin{equation}
\hat{f}(v_j | \{v_k\}) = \frac{1}{|\hat{T}| b} \sum_{\mathbf{t}_i \in \hat{T}} \frac{\exp\left( -\frac{(v_j - v_{ij})^2}{2b^2} \right)}{\sqrt{2\pi}},
\end{equation}
where \( b > 0 \) is the bandwidth.

\paragraph{Key Assumptions}
\begin{itemize}
\item[A1.] The real conditional density \( f(v_j | \{v_k\}) \) is twice continuously differentiable.
\item[A2.] The bandwidth \( b \to 0 \) and \( |\hat{T}| b \to \infty \) as \( n \to \infty \).
\item[A3.] The fuzzy matching tolerance \( \epsilon \to 0 \) such that \( \hat{T} \) asymptotically covers the real conditional support.
\end{itemize}

\paragraph{Bias-Variance Decomposition}
The pointwise MSE is:
 \begin{equation}
 \begin{aligned}
\text{MSE} &= \mathbb{E}\left[ \left( \hat{f}(v_j | \{v_k\}) - f(v_j | \{v_k\}) \right)^2 \right]\\
&= \text{Bias}^2 + \text{Var}.
 \end{aligned}
 \end{equation}

\paragraph{Bias Analysis}
Using Taylor expansion~\citep{10.5555/1162264}, we have:
\begin{align}
\text{Bias} &= \mathbb{E}[\hat{f}(v_j | \{v_k\})] - f(v_j | \{v_k\}) \\
&= \frac{b^2}{2} \frac{\partial^2 f}{\partial v_j^2}(v_j | \{v_k\}) \int u^2 K(u) du + o(b^2) \\
&= \frac{b^2}{2} \frac{\partial^2 f}{\partial v_j^2}(v_j | \{v_k\}) + o(b^2).
\end{align}

\paragraph{Variance Analysis}
\begin{align}
\text{Var} &= \mathbb{E}\left[ \hat{f}(v_j | \{v_k\})^2 \right] - \left( \mathbb{E}[\hat{f}(v_j | \{v_k\})] \right)^2 \\
&= \frac{1}{|\hat{T}| b} \left( \int K(u)^2 du \right) f(v_j | \{v_k\}) \\
&+o\left( \frac{1}{|\hat{T}| b} \right) \\
&= \frac{1}{|\hat{T}| b \sqrt{4\pi}} f(v_j | \{v_k\}) + o\left( \frac{1}{|\hat{T}| b} \right).
\end{align}

\paragraph{Consistency Proof}

Under assumptions A1-A3, the Gaussian KDE estimator is asymptotically consistent:
 \begin{equation}
\lim_{n \to \infty} \mathbb{E}\left[ \left| \hat{f}(v_j | \{v_k\}) - f(v_j | \{v_k\}) \right|^2 \right] = 0.
 \end{equation}

1. \textbf{Bias Convergence:}
 \begin{equation}
|\text{Bias}| \leq C_1 b^2 + o(b^2) \to 0 \quad \text{as } b \to 0.
 \end{equation}

2. \textbf{Variance Convergence:}
 \begin{equation}
 \begin{aligned}
\text{Var} &\leq \frac{C_2}{|\hat{T}| b} + o\left( \frac{1}{|\hat{T}| b} \right) \to 0 \quad \\
&\text{if } |\hat{T}| b \to \infty.
 \end{aligned}
 \end{equation}

3. \textbf{MSE Dominance:}
Choosing \( b \sim |\hat{T}|^{-1/5} \) yields:
 \begin{equation}
\text{MSE} = O\left( |\hat{T}|^{-4/5} \right) \to 0 \quad \text{as } |\hat{T}| \to \infty.
 \end{equation}

\subsection{Conditional Likelihood Lower Bound Analysis for Conditional Normalizing Flows}

\begin{figure*}[t]
    \centering
    \begin{subfigure}[b]{0.9\linewidth}
        \centering
        \includegraphics[width=\linewidth]{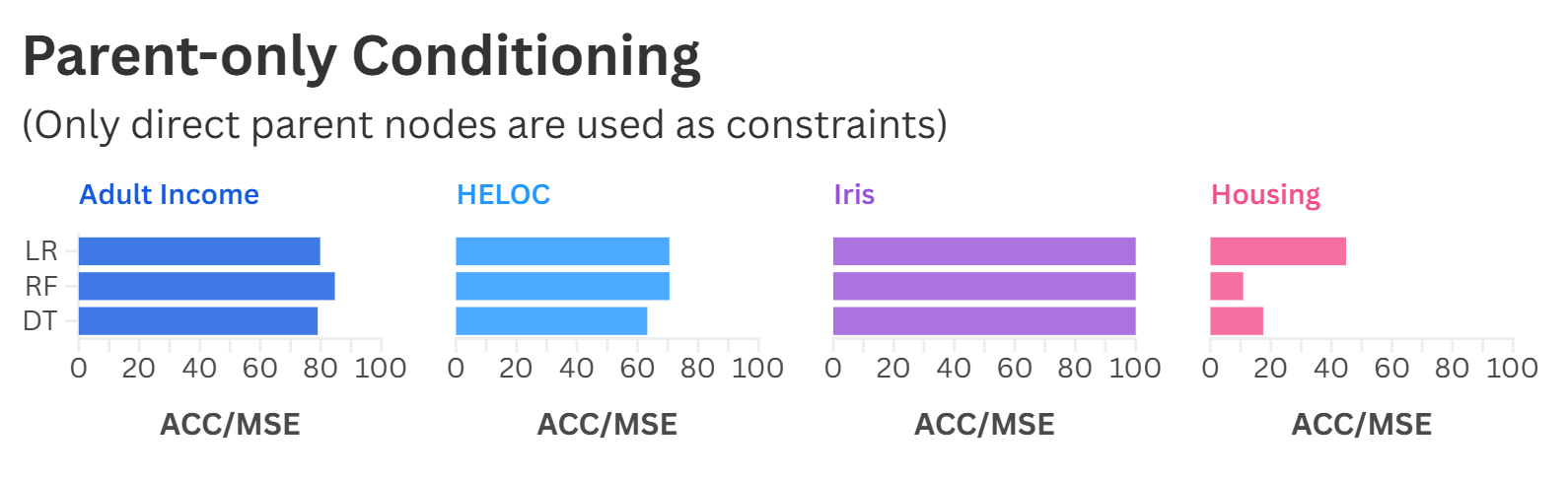}
    \end{subfigure}

    \vspace{1em}

    \begin{subfigure}[b]{0.9\linewidth}
        \centering
        \includegraphics[width=\linewidth]{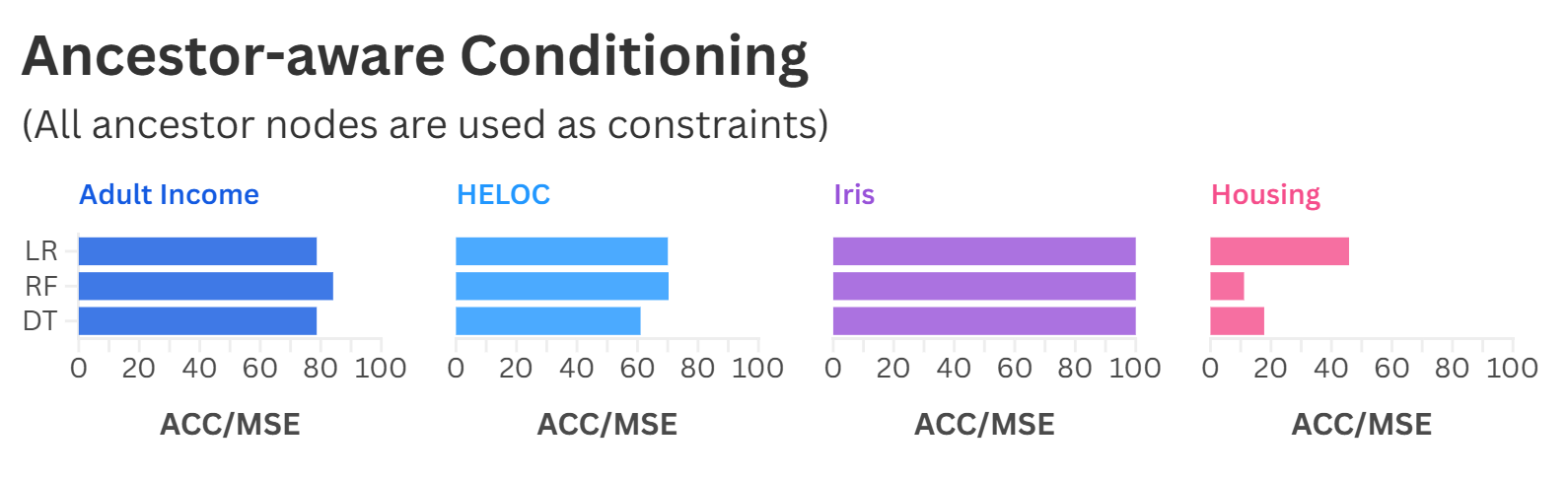}
    \end{subfigure}
    
    \caption{Comparison of synthesis performance using parent-only conditioning vs. ancestor-aware conditioning strategies. To facilitate comparison, we multiplied the MSE values by a factor of 10.}
    \label{fig:parent}
\end{figure*}

\paragraph{Problem Setup}
We model the conditional distribution $p(v_j | \{v_k : f_k \in \hat{F}_{f_j}\})$ via Conditional NFs. Let $z \sim \mathcal{N}(0, I)$ be the latent variable, and $f_\theta: \mathbb{R} \times \mathbb{R}^{|\hat{F}_{f_j}|} \to \mathbb{R}$ be the invertible transformation conditioned on parent feature values.

\paragraph{Change of Variables Theorem}
Following Eq.~\eqref{eq:log_likelihood}, the log-likelihood is given by:
\begin{equation}
\log p(v_j | \{v_k\}) = \log p_Z(z) + \log \left| \det \frac{\partial f_\theta^{-1}}{\partial v_j} \right|
\end{equation}
where $z = f_\theta^{-1}(v_j | \{v_k\})$.

\paragraph{Lipschitz-Constrained Transformation}
To ensure a stable gradient propagation, we constrain each layer $f_\theta^{(i)}$ in our flow to be $L$-Lipschitz continuous~\citep{NEURIPS2018_91d0dbfd}:
\begin{equation}
\begin{aligned}
\|f_\theta^{(i)}(z^{(i-1)} | \{v_k\}) - f_\theta^{(i)}(z'^{(i-1)} | \{v_k\})\| \\
\leq L \|z^{(i-1)} - z'^{(i-1)}\|
\end{aligned}
\end{equation}

\paragraph{Lower Bound Derivation}
1. \textbf{Layer-wise Jacobian Bound:}
For each layer, the Lipschitz continuity~\citep{kingma2018glow} implies:
\begin{equation}
\left| \det \frac{\partial f_\theta^{(i)}}{\partial z^{(i-1)}} \right| \geq L_i^{-d}
\end{equation}
where \( d \) is the dimension of \( z^{(i-1)} \).

2. \textbf{Recursive Likelihood Decomposition:}
\begin{align}
\log p(v_j | \{v_k\}) &= \log p_Z(z_0) + \\
&\sum_{i=1}^K \log \left| \det \frac{\partial f_\theta^{(i)-1}}{\partial z^{(i)}} \right| \\
&\geq \log p_Z(z_0) - \sum_{i=1}^K d \log L_i
\end{align}

3. \textbf{Equality Condition:}
Here, the bound becomes tight when each layer achieves exact Lipschitz constant \( L_i = 1 \).

\paragraph{Stability Analysis}
The gradient of the loss $\mathcal{L}(\theta) = -\log p(v_j | \{v_k\})$ satisfies:
\begin{equation}
\|\nabla_\theta \mathcal{L}\| \leq \sqrt{K} \cdot \max_{1 \leq i \leq K} \left( \frac{\|J_i\|_F}{L_i} \right)
\end{equation}
where $J_i$ is the Jacobian of layer $f_\theta^{(i)}$.

Using the Lipschitz property and the chain rule:
\begin{equation}
\begin{aligned}
\|\nabla_\theta \mathcal{L}\| &\leq \sum_{i=1}^K \left\| \frac{\partial \mathcal{L}}{\partial f_\theta^{(i)}} \right\| \cdot \| \nabla_\theta f_\theta^{(i)} \| \\
&\leq \sqrt{K} \cdot \max_i \left( \frac{\|J_i\|_F}{L_i} \right)
\end{aligned}
 \end{equation}

\paragraph{Implementation Consistency}
\begin{itemize}
\item \textbf{SwiGLU Parameterization:} The SwiGLU activation $\sigma(xW) \odot xV$ naturally satisfies $L$-Lipschitz continuity with $L = \|W\| \|V\|$
\end{itemize}

\begin{table*}[ht]
\centering
\renewcommand{\arraystretch}{1.2}
\footnotesize
\begin{tabular}{lcccccccccc}
\toprule
   & \multicolumn{2}{c}{TabSyn} & \multicolumn{2}{c}{Gemini 2.0 Flash} & \multicolumn{2}{c}{Deepseek-R1} & \multicolumn{2}{c}{Claude 3.7 Sonnet} & \multicolumn{2}{c}{GPT-4o} \\
   \cmidrule(lr){2-3} \cmidrule(lr){4-5} \cmidrule(lr){6-7} \cmidrule(lr){8-9} \cmidrule(lr){10-11}
   & ACC       & F1             & ACC                & F1              & ACC               & F1          & ACC                 & F1              & ACC          & F1          \\ \midrule
DT ($\uparrow$)& 77.58\%   & 0.70           & 78.37\%            & 0.71            & 77.14\%           & 0.69        & \textcolor{violet}{\textbf{80.13\%}}    & \textcolor{violet}{\textbf{0.73}}   & \underline{78.95}  & \underline{0.71}  \\
RF ($\uparrow$)& 83.67\%   & 0.77           & 84.99\%            & 0.77            & \underline{85.05\%}     & \underline{0.77}  & \textcolor{violet}{\textbf{85.13\%}}    & \textcolor{violet}{\textbf{0.78}}   & 84.14        & 0.75        \\
LR ($\uparrow$)& 80.21\%   & \textcolor{violet}{\textbf{0.70}}  & 80.05\%            & 0.65            & \textcolor{violet}{\textbf{80.43\%}}  & 0.65        & \underline{80.31\%}       & \underline{0.67}      & 78.61        & 0.56        \\ \bottomrule
\end{tabular}
\caption{Performance of classifiers trained on synthetic data for income-level classificaton.
\textcolor{violet}{\textbf{Bold}} indicates the best performance, and \underline{underline} indicates the second-best. ``ACC'' stands for accuracy.}
\label{tab:LLM}
\vspace{-10pt}
\end{table*}

  \begin{figure}[h]
        \centering     
        \includegraphics[width=.90\linewidth]{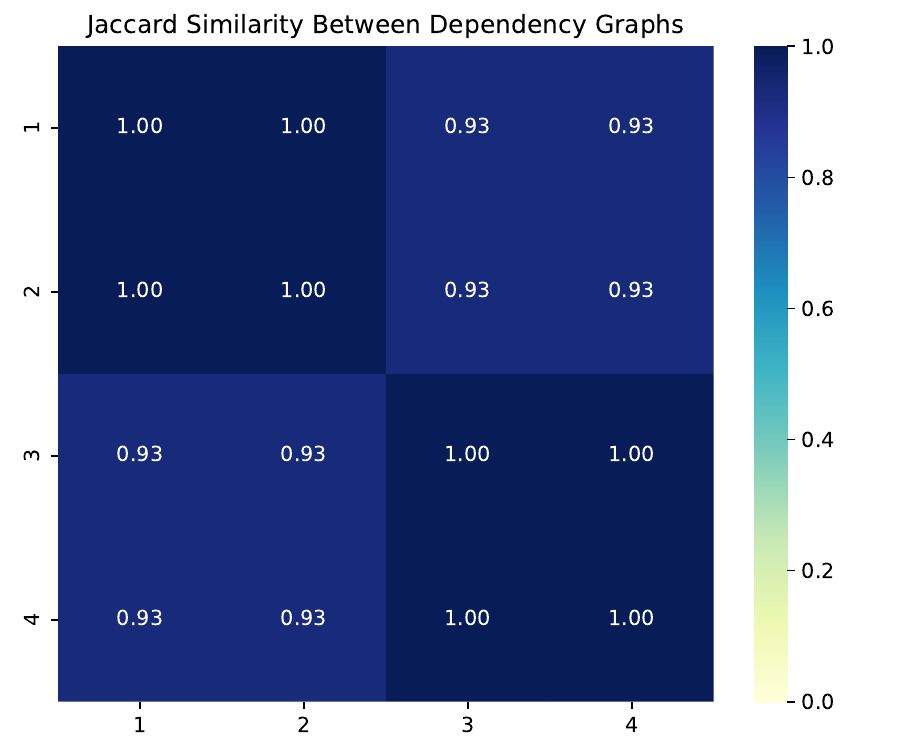}
            \caption{Heatmap of Jaccard similarities between the dependency graphs constructed from four annotation runs of the GPT-4o model. The high similarity values indicate strong consistency.}
            \label{fig:jaccard}
            \vspace{-10pt}
    \end{figure}

\begin{figure*}[t]
    \centering
    \begin{subfigure}[b]{0.24\linewidth}
        \centering
        \includegraphics[width=\linewidth]{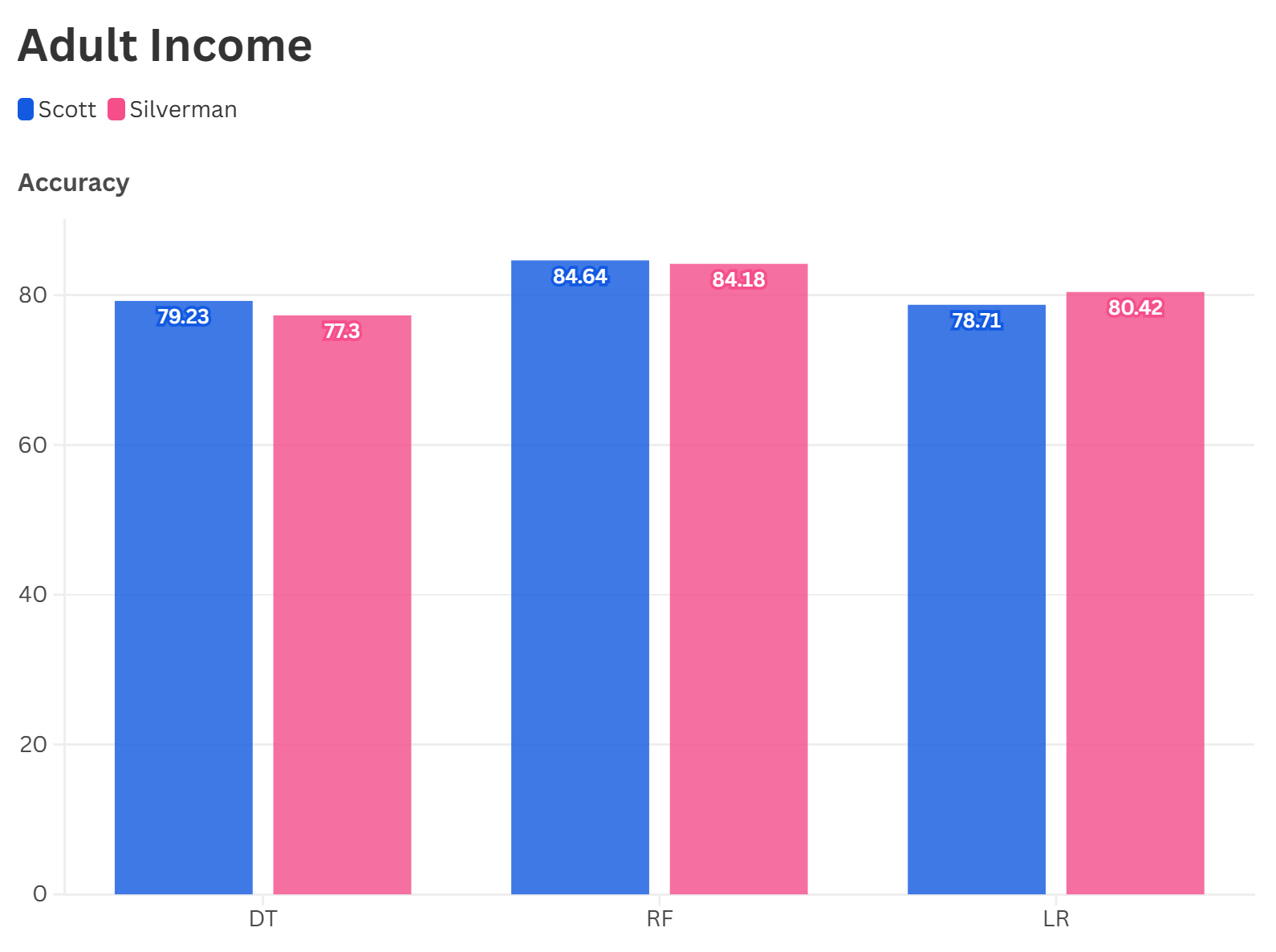}
    \end{subfigure}
    \begin{subfigure}[b]{0.24\linewidth}
        \centering
        \includegraphics[width=\linewidth]{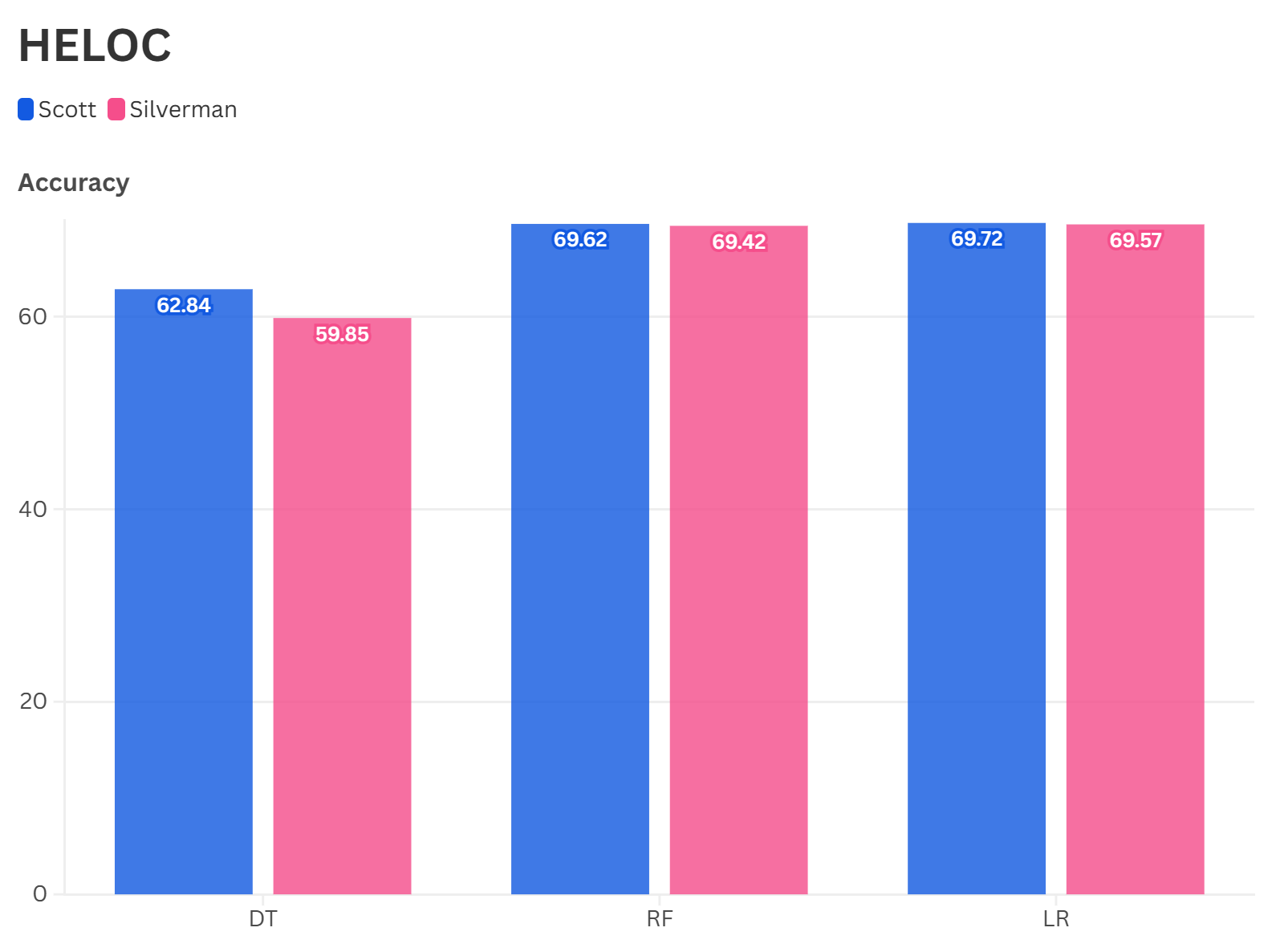}
    \end{subfigure}
        \begin{subfigure}[b]{0.24\linewidth}
        \centering
        \includegraphics[width=\linewidth]{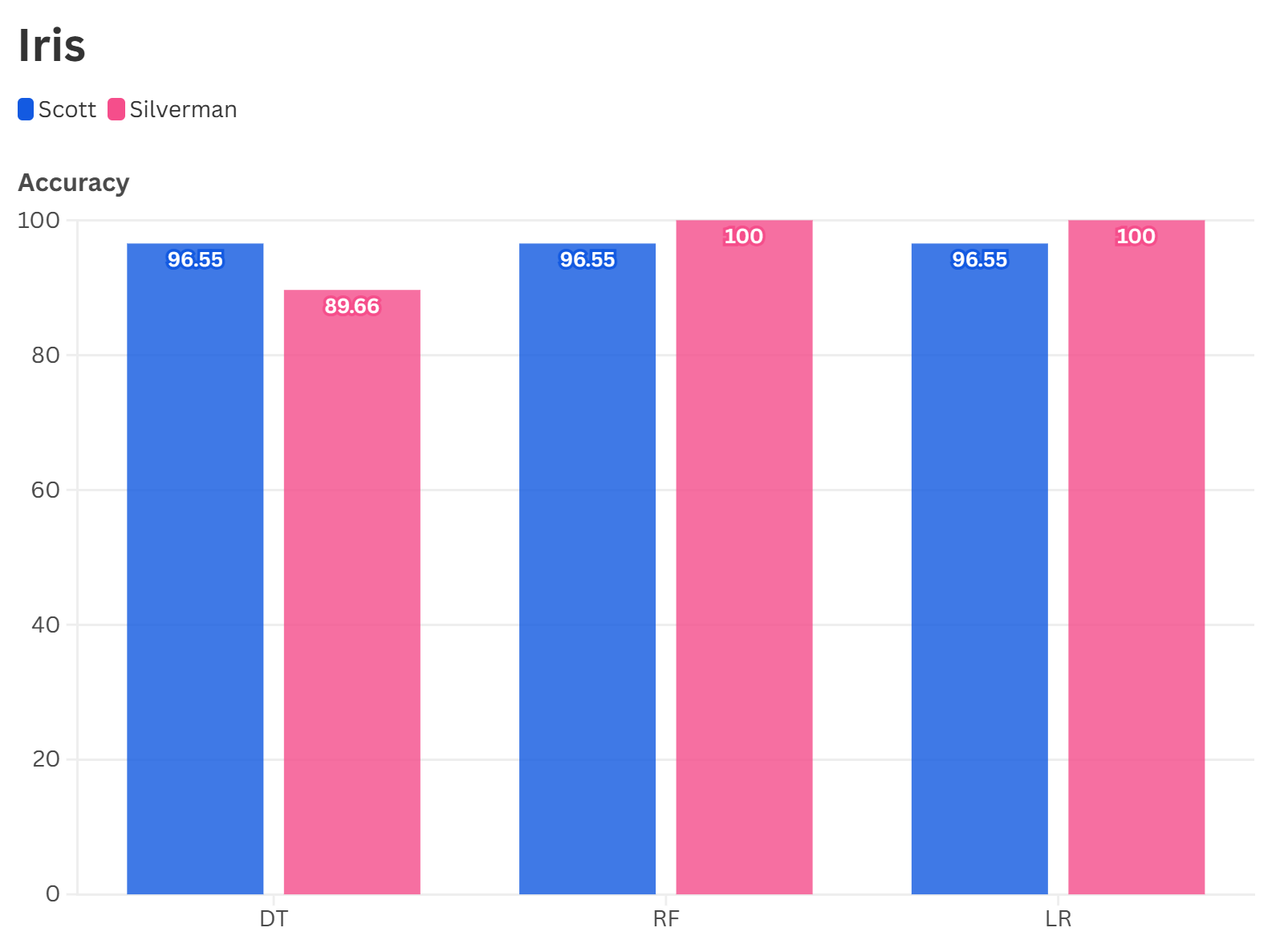}
    \end{subfigure}
        \begin{subfigure}[b]{0.24\linewidth}
        \centering
        \includegraphics[width=\linewidth]{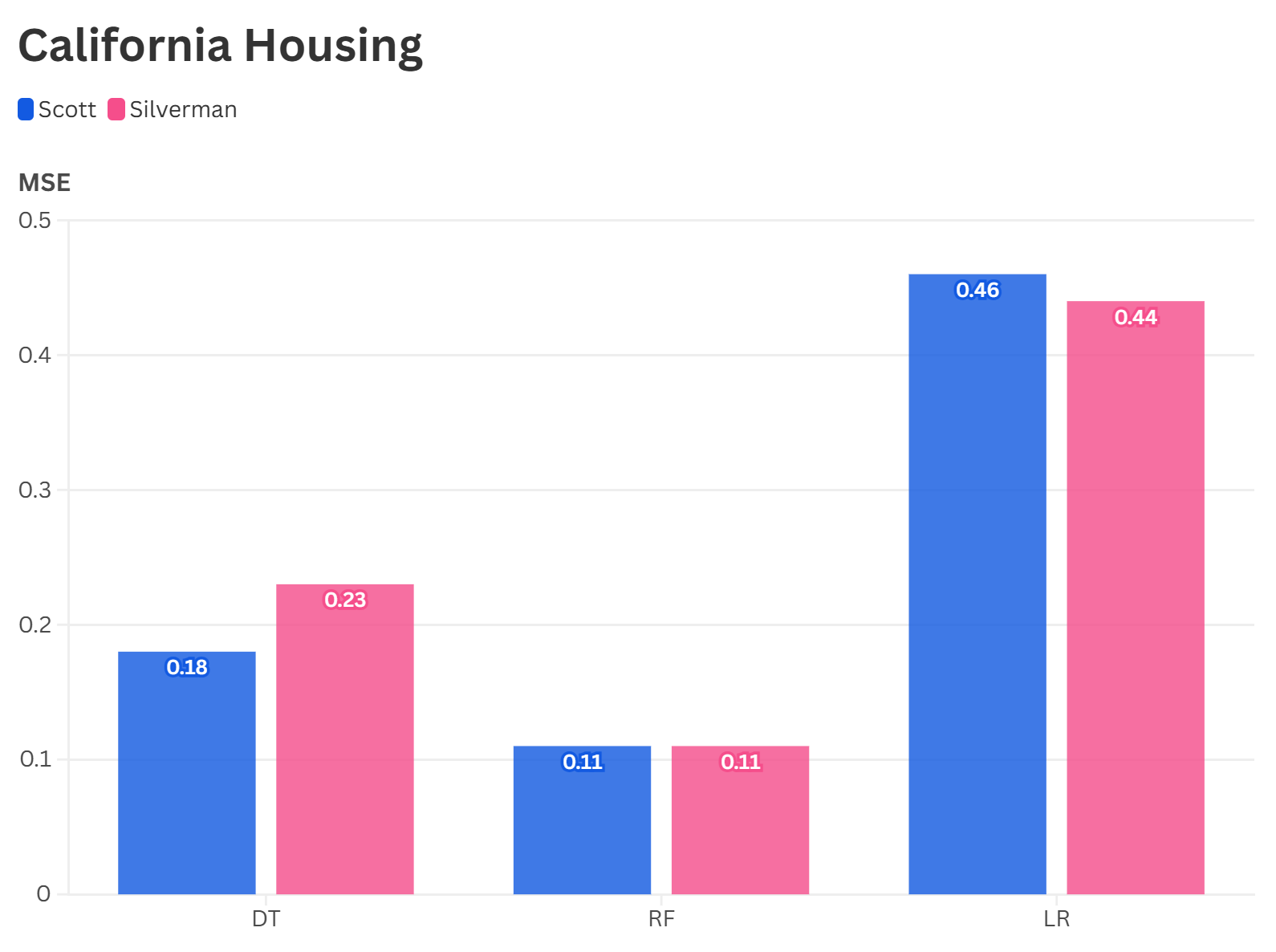}
    \end{subfigure}
    \caption{Impact of Bandwidth Selection on KDE-based Synthesis: \textcolor{blue}{Scott's rule} vs. \textcolor{magenta}{Silverman's rule}.}
    \label{fig:bandwidth}
    \vspace{-5pt}
\end{figure*}

\subsection{Time Complexity Optimization by Using BallTrees}
\label{app:balltree}
Compared to the original brute-force search, the BallTree significantly improves efficiency.
Given a dependency set $\hat{F}_{f_j}$ of size $d = |\hat{F}_{f_j}|$, the time complexity of brute-force search is:
\begin{equation}
\mathcal{O}(N \cdot d).
\end{equation}

In contrast, BallTrees reduce the average-case complexity of approximate nearest neighbor search to:
\begin{equation}
\mathcal{O}(\log N + d),
\end{equation}
under the assumption that the number of dependencies $d$ is much smaller than dataset size $N$.

\begin{table*}[ht]
\centering
\renewcommand{\arraystretch}{1.25}
\footnotesize
\resizebox{\textwidth}{!}{%
\begin{tabular}{lcccc>{\columncolor{gray!15}}c>{\columncolor{gray!15}}c}
\toprule
Dataset & \multicolumn{1}{c}{TVAE} & \multicolumn{1}{c}{CTGAN} & \multicolumn{1}{c}{GReaT} & \multicolumn{1}{c}{TabSyn} & \multicolumn{1}{c}{\textbf{Ours (w/KDE)}} & \multicolumn{1}{c}{\textbf{Ours (w/NF)}} \\ \midrule
Income ($\downarrow$)  & 77.62 $\pm$ 0.63\%                  & \underline{63.88 $\pm$ 0.49\%}             & 100\%                    & \textcolor{violet}{\textbf{52.74 $\pm$ 0.58\%}}           & 75.96 $\pm$ 0.65\%                          & 67.05 $\pm$ 0.70\%                         \\
HELOC ($\downarrow$)   & 74.33 $\pm$ 1.20\%                  & 70.49 $\pm$ 6.81\%                   & 67.91 $\pm$ 1.13\%                   & \textcolor{violet}{\textbf{52.59 $\pm$ 1.22\%}}           & \underline{58.30 $\pm$ 0.72\%}                    & 66.38 $\pm$ 0.83\%                         \\
Iris ($\downarrow$)    & 83.33 $\pm$ 7.97\%                  & 82.92 $\pm$ 10.72\%                   & 68.33 $\pm$ 8.06\%                   & \textcolor{violet}{\textbf{52.08 $\pm$ 3.17\%}}           & \underline{65.42 $\pm$ 7.03\%}                    & 65.52 $\pm$ 7.68\%                         \\
Housing ($\downarrow$) & 59.65 $\pm$ 1.60\%                  & 62.70 $\pm$ 1.07\%                   & \underline{58.14 $\pm$ 1.35\%}             & \textcolor{violet}{\textbf{50.48 $\pm$ 0.48\%}}           & 76.75 $\pm$ 1.15\%                          & 68.33 $\pm$ 0.95\%                         \\ \midrule
 Mean ($\downarrow$)    & 73.73                  & 69.99                   & 73.59                  & \textcolor{violet}{\textbf{51.97}}           & 69.05                          & \underline{68.92}                   \\ \bottomrule
\end{tabular}}
\caption{Discriminator measure with a 5-fold cross-validation. Lower accuracy values indicate that the discriminator struggles to distinguish synthetic records from real data.}
\label{tab:realism}
\vspace{-10pt}
\end{table*}

\section{Ablation Study}
To further demonstrate the effectiveness of our proposed methods, we conduct three ablation studies to answer the following questions:
\begin{enumerate}
    \item Are the LLM-based annotations robust? 
    \item Does the bandwidth significantly affect the performance of KDE-based synthesis?
    \item Is the dependency relation transitive in the generated dependency graph?
\end{enumerate}

\paragraph{Cross-LLM Annotation.} As mentioned in \S\ref{sec:limitation}, the accuracy of annotations produced by LLMs could theoretically affect the effectiveness of our approach. To assess this, we evaluated the robustness of the NL-based method by testing the downstream performance of synthetic data generated using annotations derived from different LLMs, while keeping the training and sampling settings exactly the same. We selected the Adult dataset—on which all baselines perform well according to Table~\ref{tab:downstream_utility} for this experiment. We employed Deepseek-R1~\citep{guo2025deepseek}, Gemini 2.0 Flash~\citep{team2023gemini}, and Claude 3.7 Sonnet~\citep{TheC3}, and compared them with the best-performing baseline, TabSyn. 
The results are presented in Table~\ref{tab:LLM}. \method remains effective regardless of which LLM annotation is used. Notably, the best performance was achieved using annotations from Claude 3.7 Sonnet, which outperformed other baselines across nearly all evaluation metrics. This suggests that for a given dataset, there may not be a single unique set of complete dependency relations, further confirming the robustness of \method to variations in LLM-generated annotations.

\paragraph{Repeated Annotation.} We conducted four repeated annotations using GPT-4o on the 16 mixed-type features in the Adult Income dataset. We then computed the pairwise Jaccard similarity between the resulting dependency graphs, as shown in Figure~\ref{fig:jaccard}. We observe a high degree of consistency across multiple annotations, with the first and second annotations, also the third and fourth, producing identical dependency graphs. This highlights the reliability of using LLMs for dependency annotation.


\paragraph{Scott's Rule vs Silverman's Rule.} For the KDE-based method, the bandwidth $h$ in Eq.~\eqref{eq:kde} is selected using two widely adopted empirical rules: \textbf{Scott's rule}~\citep{scott2015multivariate} and \textbf{Silverman's rule}~\citep{silverman2018density}:
\begin{equation}
h_{\text{Scott}} = n^{-1/(d+4)},
\end{equation}
\begin{equation}
h_{\text{Silverman}} = \left(\frac{4}{d+2}\right)^{1/(d+4)} n^{-1/(d+4)},
\end{equation}
where $n$ is the number of samples and $d$ is the dimensionality of the feature space.

The comparison results are shown in Figure~\ref{fig:bandwidth}. We observe that the downstream performance is similar regardless of the choice of bandwidth rule. This indicates that the effectiveness of KDE-based synthesis is not sensitive to hyperparameter selection, further supporting the robustness of \method.

\paragraph{Parent-only Conditioning vs Ancestor-aware Conditioning.} For the NF-based method, we compare two strategies for conditioning on feature dependencies: one that conditions only on direct parents, and another that conditions on all ancestors. As shown in Figure~\ref{fig:parent}, both approaches yield nearly identical downstream performance. This suggests that the constraints are not transitive, validating the precision of our dependency annotation. It further confirms the reliability of using LLMs to construct accurate dependency graphs.

\begin{figure*}[t]
    \centering

    \begin{subfigure}[b]{0.4\linewidth}
        \centering
        \includegraphics[width=\linewidth]{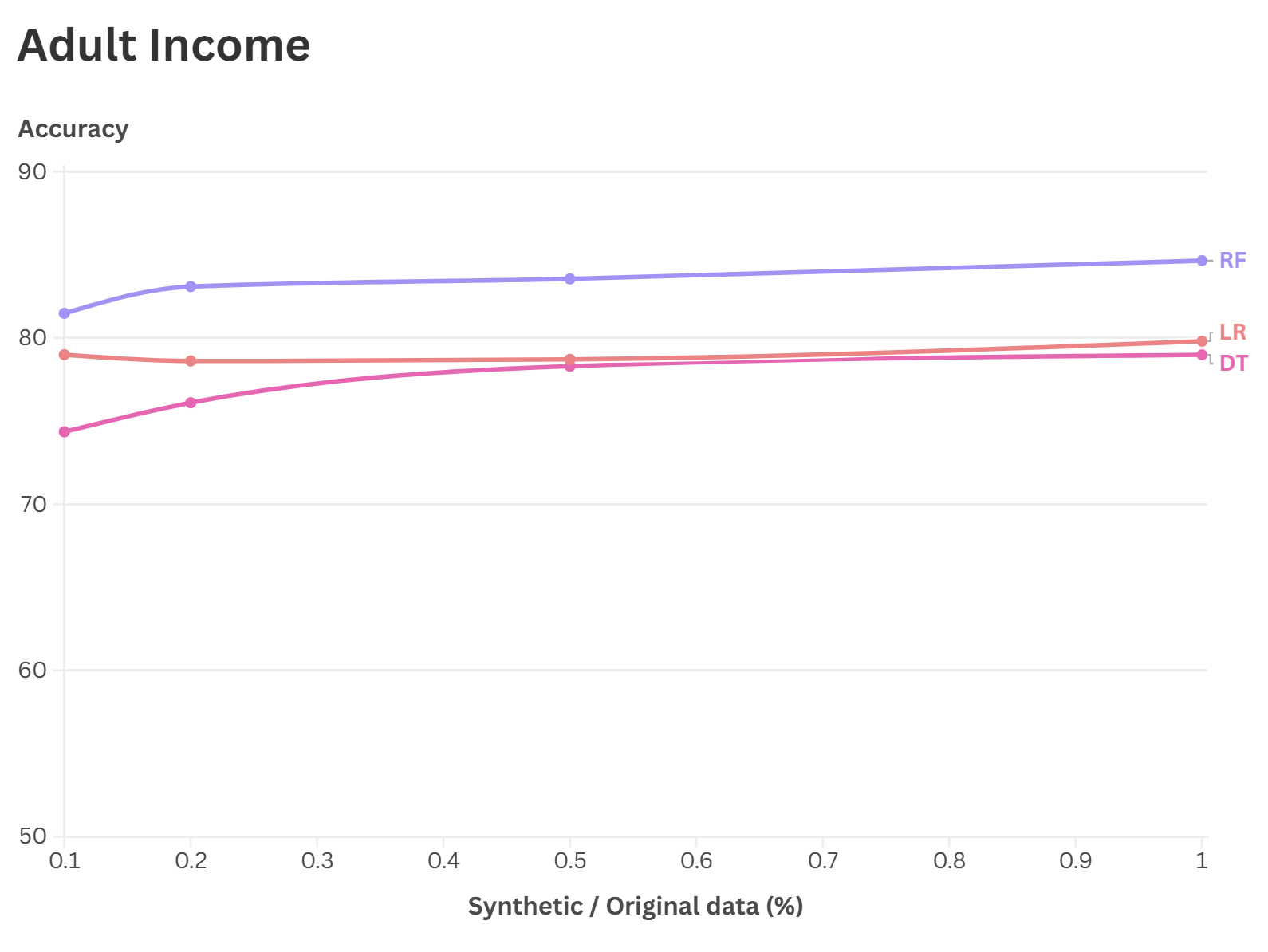}
        \caption{Adult Income}
    \end{subfigure}
    \begin{subfigure}[b]{0.4\linewidth}
        \centering
        \includegraphics[width=\linewidth]{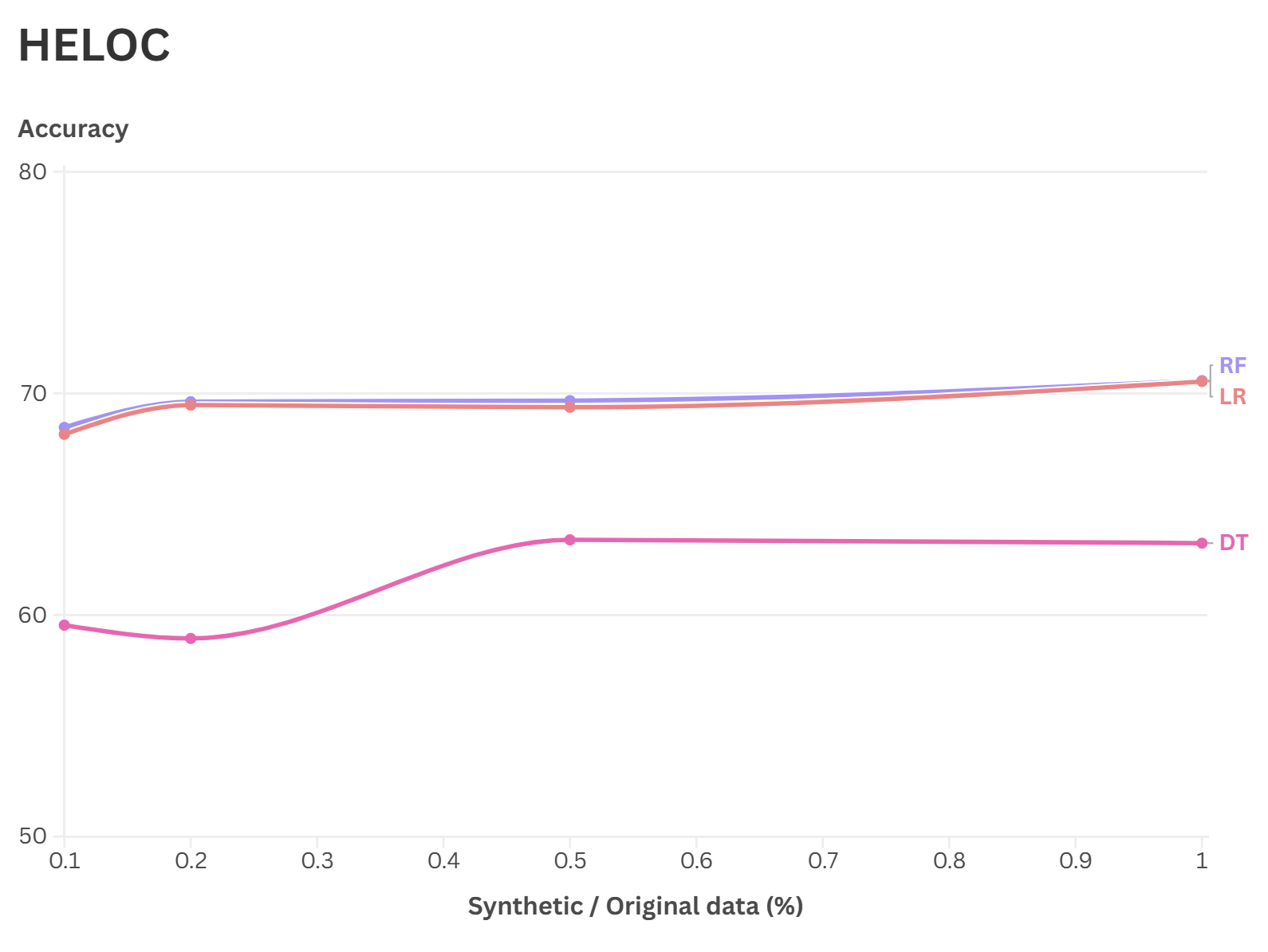}
        \caption{HELOC}
    \end{subfigure}
    
    \begin{subfigure}[b]{0.4\linewidth}
        \centering
        \includegraphics[width=\linewidth]{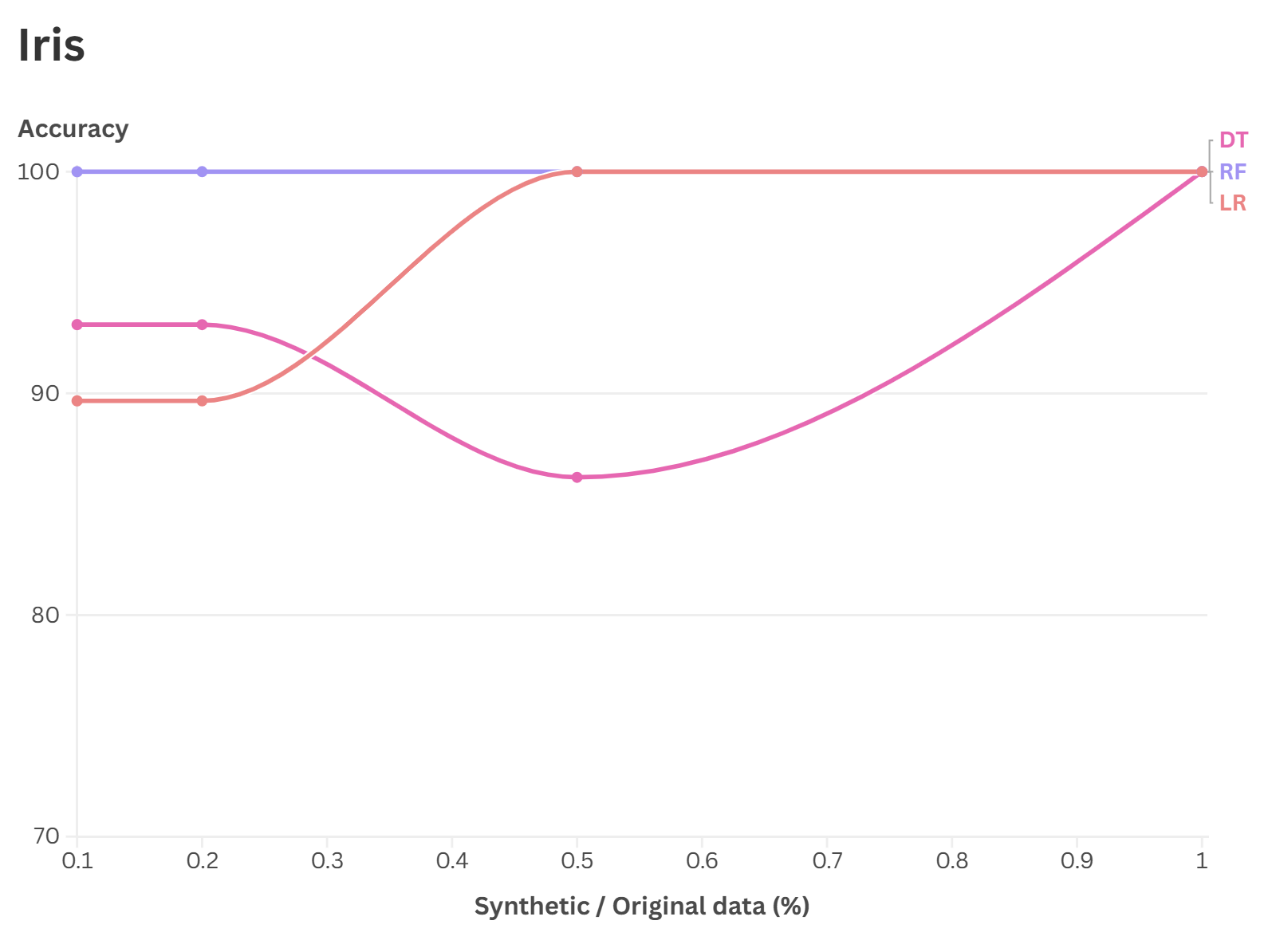}
        \caption{Iris}
    \end{subfigure}
    \begin{subfigure}[b]{0.4\linewidth}
        \centering
        \includegraphics[width=\linewidth]{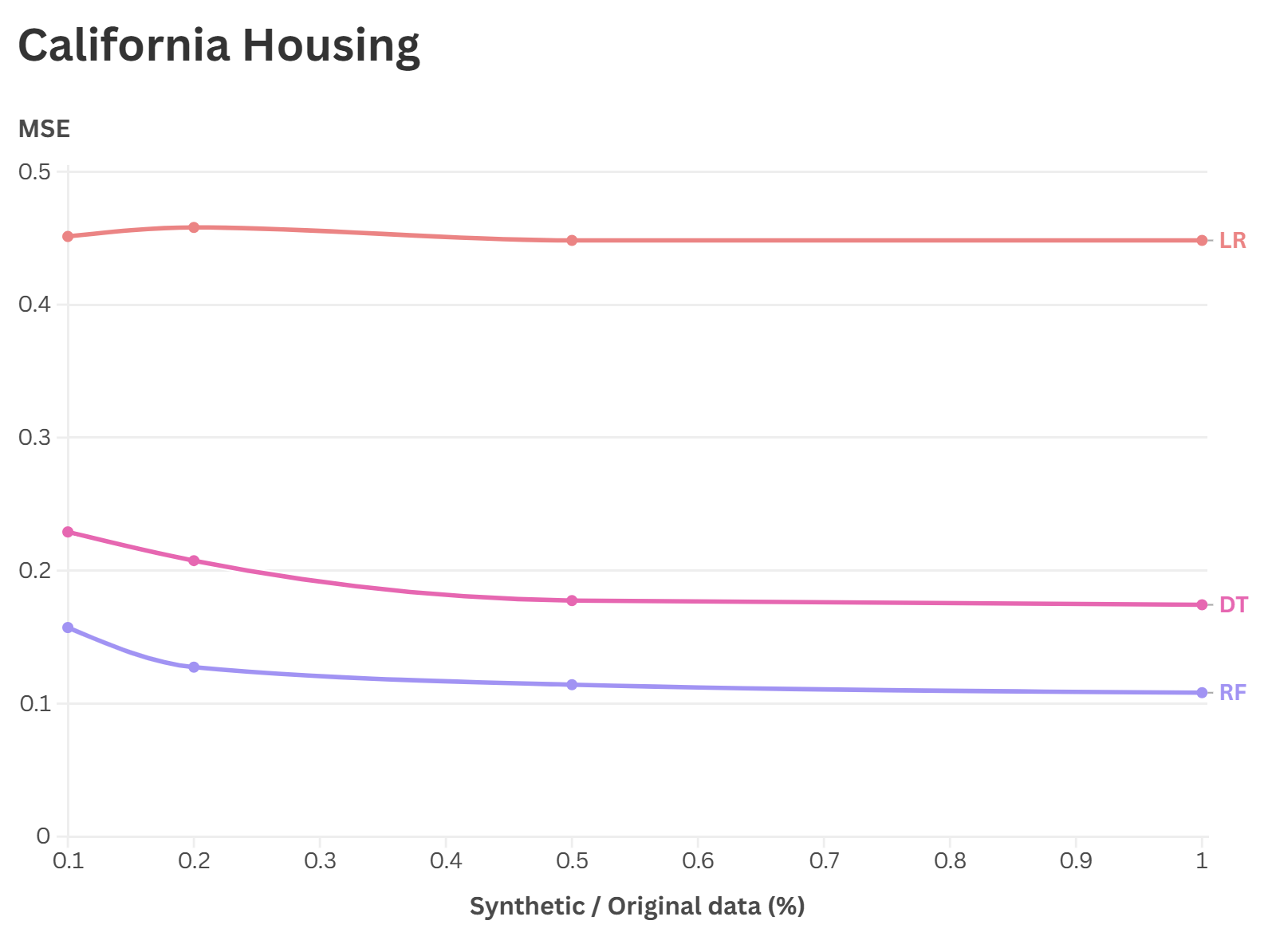}
        \caption{Housing}
    \end{subfigure}
    \caption{Model performance with increasing synthetic data (ratio to the original dataset size).}
    \label{fig:ratio}
    \vspace{-10pt}
\end{figure*}

\begin{figure*}[t]
    \centering
    \begin{subfigure}[b]{0.45\linewidth}
        \centering
        \includegraphics[width=\linewidth]{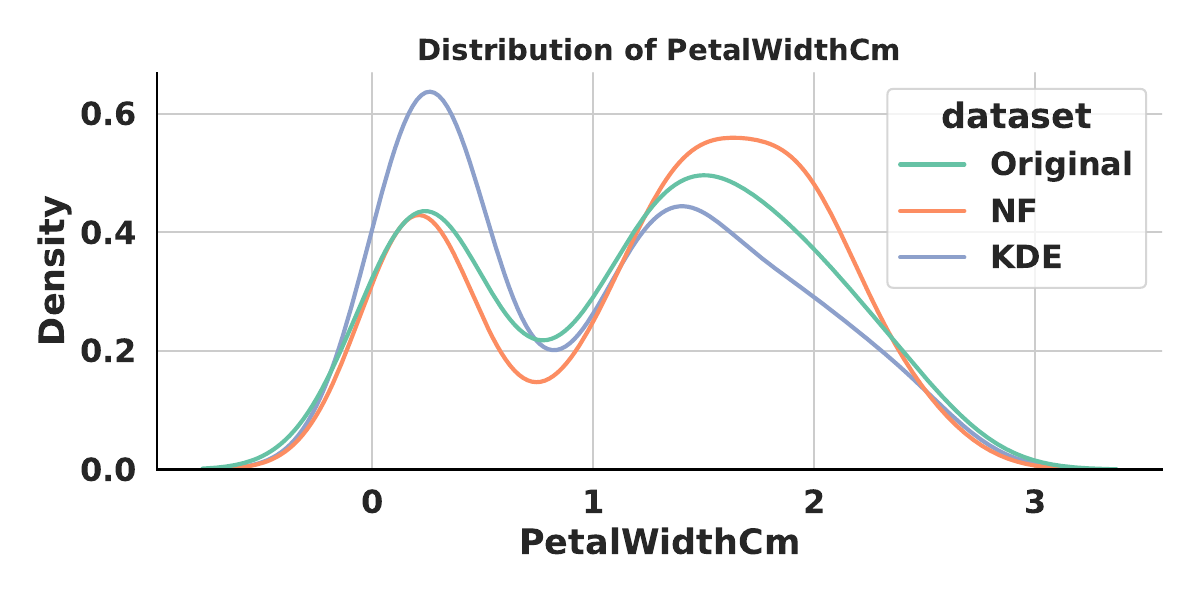}
    \end{subfigure}
    \begin{subfigure}[b]{0.45\linewidth}
        \centering
        \includegraphics[width=\linewidth]{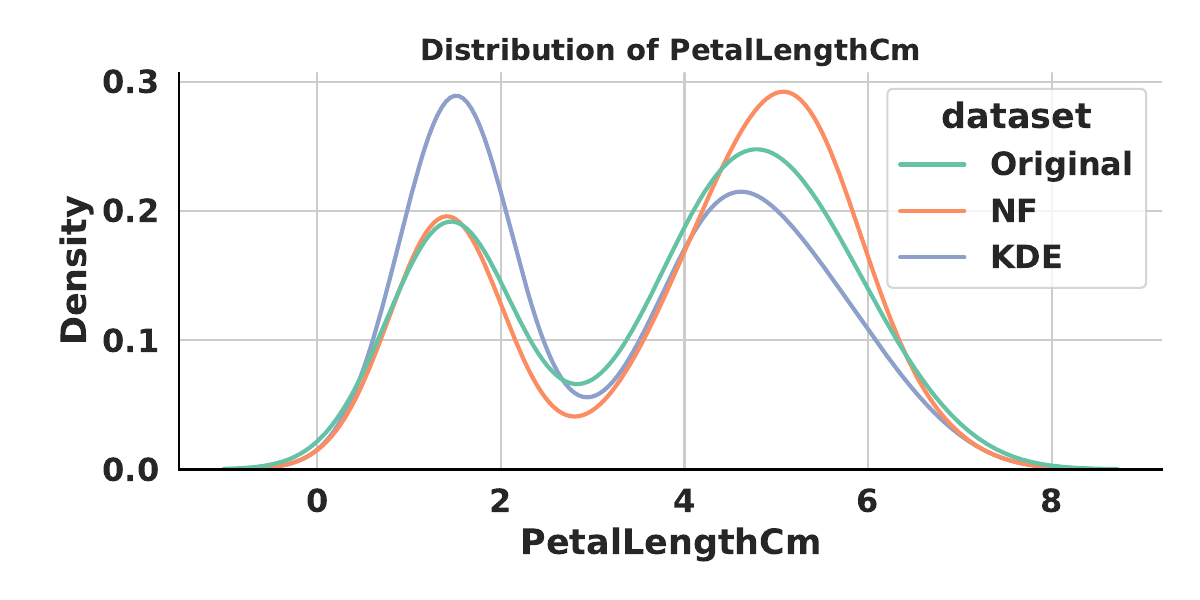}
    \end{subfigure}
    
    \vspace{1em}

        \begin{subfigure}[b]{0.45\linewidth}
        \centering
        \includegraphics[width=\linewidth]{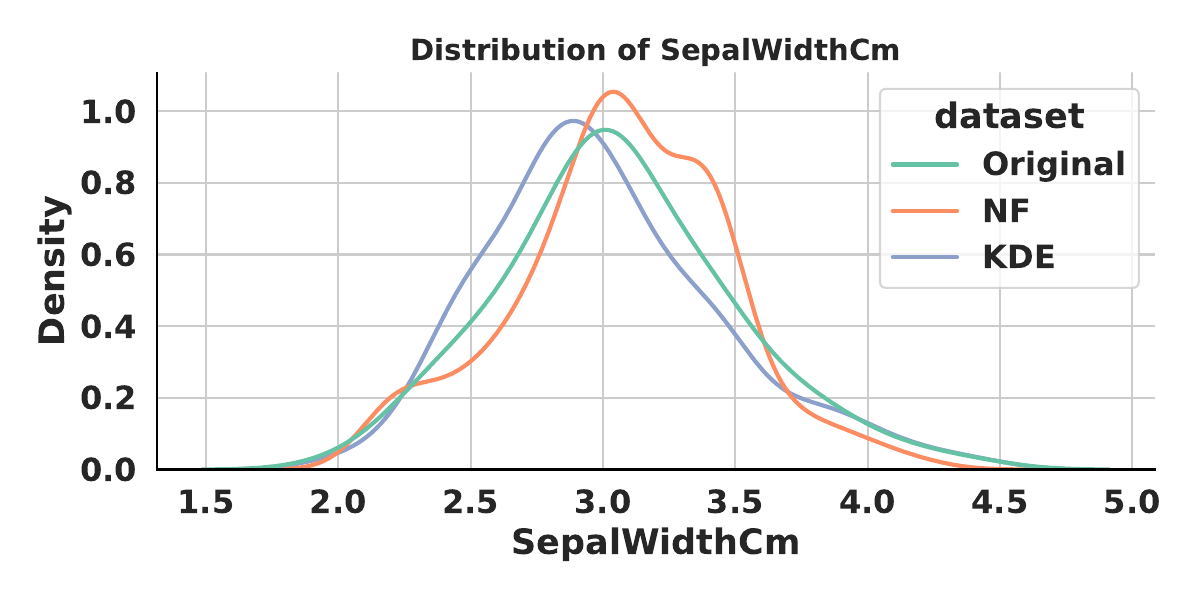}
    \end{subfigure}
    \begin{subfigure}[b]{0.45\linewidth}
        \centering
        \includegraphics[width=\linewidth]{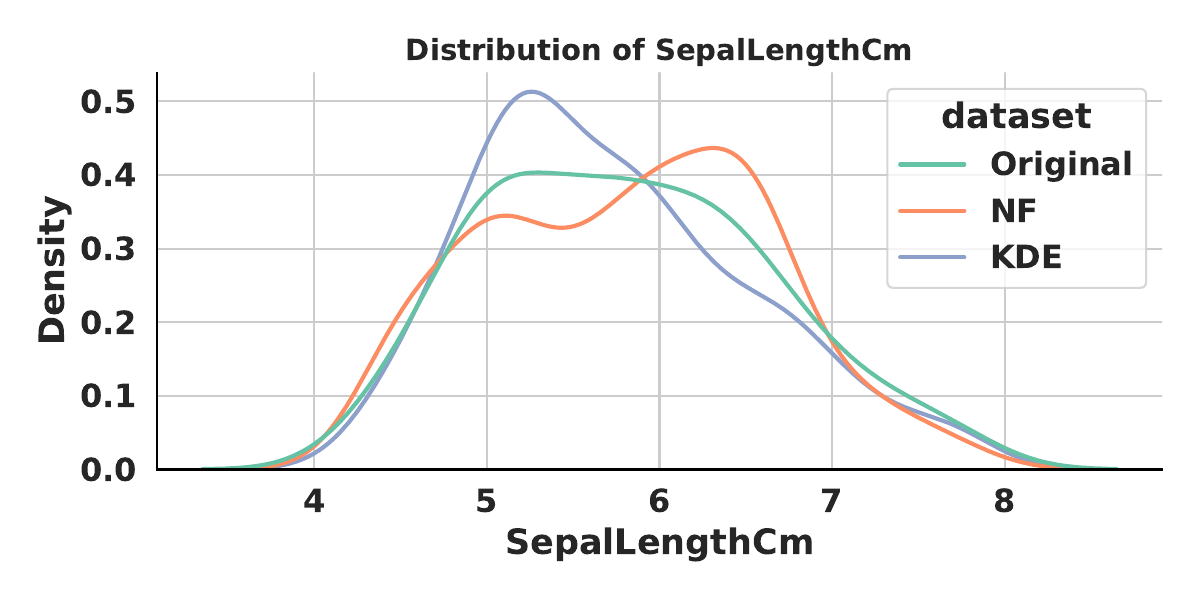}
    \end{subfigure}
    \caption{A visualization of the density distributions of Sepal and Petal lengths and widths on the Iris dataset, comparing the original and synthetic data.}
    \label{fig:iris}
\end{figure*}

\begin{table*}[ht] 
    \centering
    \resizebox{\textwidth}{!}{
    \begin{tabular}{lccccccc}
            \toprule
            Dataset  & Domain & \#Samples & \#Features & Task   & \#Classes\\
            \midrule
            Income \citep{adult_2}  & Social     & 48,842   & 15     & Classification    & 2  \\
            HELOC \citep{heloc}  & Finance     & 10,459     & 24             & Classification & 2\\
            Iris \citep{iris_53}   & Biology      & 150   & 5         & Classification & 3 \\
            Housing \citep{house} & Real Estate     & 20,640      & 10           & Regression & N/A \\
            CDC \citep{burrows2017incidence}  & Healthcare & 253,680   & 20        & Classification     & 3  \\
            Mushroom \citep{Wagner2021MushroomDC}  & Biology & 61,068   & 19        & Classification     & 2 \\
        \bottomrule
        \end{tabular}}
  \caption{The statistics of the datasets employed in our experiments.} 
  \label{tab:datasets}
\end{table*}

\section{More Experiment}

\paragraph{Realism.} 
We trained a support vector machine~\citep{svm} using 5-fold cross-validation to distinguish between the original dataset $T$ and the synthetic dataset $T'$. The accuracy serves as the discriminator measure. High-quality synthetic data should be difficult for the discriminator to distinguish from real data. The results are shown in Table~\ref{tab:realism}.

Based on Table~\ref{tab:realism}, we observe that \method produces synthetic data that substantially confuses classifiers trained on real data, indicating a high degree of realism. Although our approach underperforms TabSyn in terms of the average discriminator accuracy, it still demonstrates an advantage over the other baselines. Theoretically, for perfectly realistic data, the discriminator accuracy should approach 50\%. Among all methods, only TabSyn, which is based on diffusion models, achieves this ideal. We attribute this to TabSyn’s use of a score function in the latent space, which guides the generation process toward high-probability regions and thus produces samples closely aligned with the original data distribution. However, this close resemblance may also result in considerable feature overlap between real and synthetic samples, potentially raising privacy concerns.

\paragraph{Synthetic Data Size.}
To demonstrate that the classifier/regressors used in our evaluation have been sufficiently trained on our augmented data, we assessed the impact of varying augmentation sizes on model performance. 

As shown in Figure~\ref{fig:ratio}, we observe that the performance of nearly all classifier/regressors converges to their respective optima when the size of synthetic data reaches approximately 0.5 times that of the original dataset. This finding supports that all evaluation models in our experiments have been adequately trained, further validating the soundness and reproducibility of our experimental setup.

\paragraph{Visualization.} 
To compare the distributional differences between our synthetic data and the original data, we visualize the density distributions of the four numerical features used to determine the iris species in the Iris dataset—namely, the lengths and widths of the Sepal and Petal. As shown in Figure~\ref{fig:iris}, we observe that both of our methods are able to accurately capture the distributional patterns of the original data.

\section{Reproducibility}
\subsection{Hardware Environment}
\label{sec:hardware}
The experimental hardware environment we used is shown in Table~\ref{tab:hardware}.
    \begin{table}[h]
    \centering
    \footnotesize
    \renewcommand{\arraystretch}{1.1}
    \begin{tabular}{lc}
    \toprule
    Memory            & 1012G                  \\
    CPU               & AMD EPYC 7763 2.45G Hz \\
    GPU              & 4 x NVIDIA A100 80G\\
    Operating        &Ubuntu \\
    System           & 20.04.6 LTS     \\ \bottomrule
    \end{tabular}
    \caption{Experimental hardware environment.}
    \label{tab:hardware}
    \vspace{-10pt}
    \end{table} 
    
\subsection{Training Details}
In our experiments, the conditional batch normalization module is implemented using a single linear layer, a single SwiGLU activation layer, and a layer normalization module. The linear layer contains 128 neurons, with a dropout rate of 0.1. During training, we employ batch gradient descent with the AdamW optimizer~\citep{loshchilov2017decoupled}, and the learning rate is set to $5 \times 10^{-4}$. The initial input distribution for the normalizing flow NF is a standard normal distribution, and the kernel density estimation is based on a standard Gaussian kernel.
The statistics of the datasets used is shown in Table~\ref{tab:datasets}.

\section{Scalability Analysis}  
The scalability of tabular augmentation methods may become crucial when dealing with high-dimensional datasets. In domains such as genomics~\citep{kelleher2013processing}, the feature counts are extremely high in theory. Our framework inherently addresses this challenge through two key designs: \textbf{sparse dependency modeling} and \textbf{modular generation pipelines}.  

Traditional LLM-based methods incur quadratic memory overhead in self-attention layers, making them impractical for datasets with excessive features. In contrast, our sparse dependency extraction reduces pairwise interactions to \(O(M \cdot k)\), where \(k \ll M\) is the average number of parent nodes per feature (empirically \(k \leq 6\) across our four datasets). Topological traversal ensures linear time complexity \(O(M)\) during sampling.

\begin{table*}[t]
\centering
\renewcommand{\arraystretch}{1.1}
\footnotesize
\begin{tabular}{llcccccccc}
\toprule
                     \multirow{2}{*}{Dataset}     &    & \multicolumn{2}{c}{Original} & \multicolumn{2}{c}{TVAE} & \multicolumn{2}{c}{CTGAN} & \multicolumn{2}{c}{\textbf{Ours (w/NF)}} \\
                     \cmidrule(lr){3-4} \cmidrule(lr){5-6} \cmidrule(lr){7-8} \cmidrule(lr){9-10}
                   &    & ACC            & F1          & ACC         & F1         & ACC         & F1          & ACC                  & F1                \\ \midrule
\multirow{3}{*}{CDC ($\uparrow$)}      
                          & DT & 73.65\%        & 0.40        & \textcolor{violet}{\textbf{71.96\%}}         & \textcolor{violet}{\textbf{0.46}}        & \underline{71.51\%}         & 0.36          & 71.19\%              & \underline{0.36}              \\
                          & RF & 83.15\%        & 0.38        & 82.44\%         & \textcolor{violet}{\textbf{0.43}}        & \underline{82.67\%}         & \underline{0.35}         & \textcolor{violet}{\textbf{82.96\%}}              & 0.31              \\
                          & LR & 82.70\%        & 0.40        & 81.62\%         & \textcolor{violet}{\textbf{0.48}}        & \underline{81.98\%}         & \underline{0.36}         & \textcolor{violet}{\textbf{82.77\%}}              & 0.31              \\ \midrule
\multirow{3}{*}{Mushroom ($\uparrow$)} 
                          & DT & 52.49\%        & 0.47        & 51.30\%          & \textcolor{violet}{\textbf{0.51}}        & 51.38\%         & \underline{0.50}         & \textcolor{violet}{\textbf{51.79\%}}              & 0.45              \\
                          & RF & 59.22\%        & 0.44        & \underline{63.96\%}          & \textcolor{violet}{\textbf{0.58}}        & 60.57\%         & 0.53         & \textcolor{violet}{\textbf{64.06\%}}              & \underline{0.54}              \\
                          & LR & 55.97\%        & 0.36        & \underline{55.96\%}          & 0.35        & 55.88\%         & \underline{0.35}         & \textcolor{violet}{\textbf{55.97\%}}              & \textcolor{violet}{\textbf{0.36}}              \\ \bottomrule
\end{tabular}
\caption{Performance of classifiers trained on synthetic data for downstream tasks.}
\label{tab:large}
\end{table*}

\subsection{Memory and Storage Optimization}  
We recommend the following optimizations:
\begin{enumerate}
\item \textbf{Sparse Graph Representation}: Use adjacency lists for dependency edges, reducing memory from \(O(M^2)\) to \(O(M \cdot k)\).  
\item \textbf{Parallelized NF Training}: Deploy feature-specific normalizing flows on GPUs, sharing parameters for categorical features with similar dependencies.  
\item \textbf{On-Demand KDE Sampling}: Cache BallTree indices in distributed key-value stores for large datasets. 
\end{enumerate}

\subsection{Practical Limitations and Mitigations}  
While theoretically scalable, two bottlenecks emerge in practice:  
\begin{itemize}
\item \textbf{LLM Annotation Overhead}: Prompting LLMs for a large number of features incurs prohibitive costs. Future work may use lightweight predictors, e.g., GNN~\citep{scarselli2008graph}, to bootstrap the process.  
\item \textbf{Feature Interaction Sparsity}: Assumes local dependencies dominate, which holds empirically in real-world tabular data~\citep{liu2023goggle}. For systems with global interactions, hybrid architectures combining \method with low-rank attention layers~\citep{hu2022lora} could be explored.
\end{itemize}

\section{Experimental Evaluations on Real-world and Large-scale Datasets}
Based on the scalability discussion above, we conducted experiments on two additional large-scale datasets and reported the downstream utility under synthetic data, as shown in Table~\ref{tab:large}.
Due to the large scale of the datasets, it was impractical to compare with LLM- and diffusion-based methods. Therefore, we compared against TVAE and CTGAN.

\paragraph{CDC Diabetes Health Indicators.} The \emph{CDC} dataset~\citep{burrows2017incidence} comprises over 250,000 samples, containing healthcare statistics and lifestyle survey responses, along with diabetes diagnoses. It includes 35 features covering demographics and laboratory test results. The task is to predict whether a patient is healthy, Diabetes mellitus Typ 1, or Diabetes mellitus Typ 2.

\paragraph{Mushroom.} The \emph{Mushroom} dataset~\citep{Wagner2021MushroomDC} comprises over 61,000 samples, containing biological features of mushrooms for binary classification into edible and poisonous.

The experiments conducted on larger-scale datasets further support the conclusions drawn from Table~\ref{tab:downstream_utility}, demonstrating the scalability of \method.

\section{Prompt Used}
\label{app:prompt}

The prompt template we used is shown in Table~\ref{tab:prompt}.

\begin{table}[ht]
\centering
\begin{adjustbox}{max width=0.95\textwidth}
\begin{tcolorbox}[colback=gray!5!white, colframe=gray!50!black, 
    boxrule=0.5pt, arc=2mm, fonttitle=\bfseries, title=LLM Prompt]
\small
\begin{spacing}{1}
\noindent Given a tabular dataset with the following description: \\
``\texttt{\{description\}}'' \\

The dataset holds the following features, represented in numbers or text strings: \\
\texttt{\{numerical\_feature + categorical\_feature\}} \\

Please list the constraints for each feature based on the others. Return the results in the following format: for each feature, first output the feature name followed by a colon, and then a set of constraints represented by square brackets. The `\texttt{->}` symbol indicates that the former is the cause and the latter is the effect. Different constraints should be separated by commas. \\

Here is an example: \\
\texttt{Feature A: [Feature B->Feature A, Feature C->Feature A]} \\
This means that both Feature B and Feature C determine the range of Feature A. \\

Please leave it blank if there is no relation between a feature and others. 
\end{spacing}
\end{tcolorbox}
\end{adjustbox}
\caption{Prompt used for dependency annotation.}
\label{tab:prompt}
\vspace{-10pt}
\end{table}

\section{Dependencies Extracted}
\label{app:dependency}
The follows are the complete dependencies extracted from the datasets we used, as shown in Table~\ref{tab:dep_iris}, ~\ref{tab:dep_income}, ~\ref{tab:dep_mushroom}, ~\ref{tab:dep_housing}, ~\ref{tab:dep_heloc} and ~\ref{tab:dep_diabetes}.

\begin{table}[htbp]
\centering
\footnotesize
\renewcommand{\arraystretch}{1.1}
\begin{tabular}{p{0.22\linewidth} p{0.72\linewidth}}
\toprule
\textbf{Feature} & \textbf{Dependencies} \\
\midrule
Sepal-Length-Cm & --- \\
Sepal-Width-Cm & --- \\
Petal-Length-Cm & --- \\
Petal-Width-Cm & --- \\
Species & SepalLengthCm, SepalWidthCm, PetalLengthCm, PetalWidthCm $\rightarrow$ Species \\
\bottomrule
\end{tabular}
\caption{Extracted Feature Dependencies for \texttt{Iris} Dataset.}
\label{tab:dep_iris}
\vspace{-10pt}
\end{table}

\begin{table}[htbp]
\centering
\footnotesize
\renewcommand{\arraystretch}{1.1}
\begin{tabular}{p{0.22\linewidth} p{0.72\linewidth}}
\toprule
\textbf{Feature} & \textbf{Dependencies} \\
\midrule
age & --- \\
fnlwgt & --- \\
educational-num & education $\rightarrow$ educational-num \\
capital-gain & occupation $\rightarrow$ capital-gain \\
capital-loss & occupation $\rightarrow$ capital-loss \\
hours-per-week & occupation $\rightarrow$ hours-per-week \\
workclass & occupation, education $\rightarrow$ workclass \\
education & educational-num $\rightarrow$ education \\
marital-status & age $\rightarrow$ marital-status \\
occupation & education, workclass $\rightarrow$ occupation \\
relationship & marital-status, gender $\rightarrow$ relationship \\
race & --- \\
gender & --- \\
native-country & --- \\
income & education, workclass, occupation, capital-gain, capital-loss, hours-per-week $\rightarrow$ income \\
\bottomrule
\end{tabular}
\caption{Extracted Feature Dependencies for \texttt{Income} Dataset.}
\label{tab:dep_income}
\vspace{-10pt}
\end{table}

\begin{table}[htbp]
\centering
\footnotesize
\renewcommand{\arraystretch}{1.1}
\begin{tabular}{p{0.28\linewidth} p{0.68\linewidth}}
\toprule
\textbf{Feature} & \textbf{Dependencies} \\
\midrule
cap-diameter & --- \\
stem-height & --- \\
stem-width & --- \\
class & cap-shape, cap-surface, cap-color, does-bruise-or-bleed, gill-attachment, gill-spacing, gill-color, stem-root, stem-surface, stem-color, veil-type, veil-color, has-ring, ring-type, spore-print-color, habitat, season $\rightarrow$ class \\
cap-shape & --- \\
cap-surface & --- \\
cap-color & --- \\
does-bruise-or-bleed & --- \\
gill-attachment & --- \\
gill-spacing & --- \\
gill-color & --- \\
stem-root & --- \\
stem-surface & --- \\
stem-color & --- \\
veil-type & --- \\
veil-color & --- \\
has-ring & --- \\
ring-type & --- \\
spore-print-color & --- \\
habitat & --- \\
season & --- \\
\bottomrule
\end{tabular}
\caption{Extracted Feature Dependencies for \texttt{Mushroom} Dataset.}
\label{tab:dep_mushroom}
\vspace{-10pt}
\end{table}

\begin{table*}[htbp]
\centering
\footnotesize
\renewcommand{\arraystretch}{1.1}
\begin{tabular}{p{0.22\linewidth} p{0.72\linewidth}}
\toprule
\textbf{Feature} & \textbf{Dependencies} \\
\midrule
longitude & --- \\
latitude & longitude $\rightarrow$ latitude \\
housing\_median\_age & longitude, latitude, ocean\_proximity $\rightarrow$ housing\_median\_age \\
total\_rooms & population, households, median\_income $\rightarrow$ total\_rooms \\
total\_bedrooms & total\_rooms, population, households $\rightarrow$ total\_bedrooms \\
population & households, total\_rooms, total\_bedrooms $\rightarrow$ population \\
households & population, total\_rooms, total\_bedrooms $\rightarrow$ households \\
median\_income & longitude, latitude, ocean\_proximity $\rightarrow$ median\_income \\
median\_house\_value & median\_income, housing\_median\_age, ocean\_proximity, latitude, longitude $\rightarrow$ median\_house\_value \\
ocean\_proximity & longitude, latitude $\rightarrow$ ocean\_proximity \\
\bottomrule
\end{tabular}
\caption{Extracted Feature Dependencies for \texttt{Housing} Dataset.}
\label{tab:dep_housing}
\vspace{-10pt}
\end{table*}

\begin{table*}[htbp]
\centering
\footnotesize
\renewcommand{\arraystretch}{1.1}
\begin{tabular}{p{0.25\linewidth} p{0.68\linewidth}}
\toprule
\textbf{Feature} & \textbf{Dependencies} \\
\midrule
ExternalRiskEstimate & NumSatisfactoryTrades, PercentTradesNeverDelq, NumTotalTrades, NumTrades90Ever2DerogPubRec $\rightarrow$ ExternalRiskEstimate \\
MSinceOldestTradeOpen & --- \\
MSince-Most-Recent-Trade-Open & MSinceOldestTradeOpen $\rightarrow$ MSinceMostRecentTradeOpen \\
AverageMInFile & MSinceOldestTradeOpen, MSinceMostRecentTradeOpen $\rightarrow$ AverageMInFile \\
NumSatisfactoryTrades & NumTotalTrades, PercentTradesNeverDelq $\rightarrow$ NumSatisfactoryTrades \\
Num-Trades-60-Ever-2-Derog-Pub-Rec & NumTotalTrades $\rightarrow$ NumTrades60Ever2DerogPubRec \\
Num-Trades-90-Ever-2-Derog-Pub-Rec & NumTotalTrades $\rightarrow$ NumTrades90Ever2DerogPubRec \\
PercentTradesNeverDelq & NumSatisfactoryTrades, NumTotalTrades $\rightarrow$ PercentTradesNeverDelq \\
MSinceMostRecentDelq & MSinceMostRecentTradeOpen $\rightarrow$ MSinceMostRecentDelq \\
Max-Delq-2-Public-Rec-Last-12M & NumTrades60Ever2DerogPubRec, NumTrades90Ever2DerogPubRec $\rightarrow$ MaxDelq2PublicRecLast12M \\
MaxDelqEver & MaxDelq2PublicRecLast12M $\rightarrow$ MaxDelqEver \\
NumTotalTrades & NumSatisfactoryTrades, NumTrades60Ever2DerogPubRec, NumTrades90Ever2DerogPubRec $\rightarrow$ NumTotalTrades \\
Num-Trades-Open-in-Last-12M & MSinceMostRecentTradeOpen, NumTotalTrades $\rightarrow$ NumTradesOpeninLast12M \\
PercentInstallTrades & NumInstallTradesWBalance, NumTotalTrades $\rightarrow$ PercentInstallTrades \\
M-Since-Most-Recent-Inqexc-l7days & --- \\
NumInqLast6M & MSinceMostRecentInqexcl7days $\rightarrow$ NumInqLast6M \\
NumInqLast6Mexcl7days & NumInqLast6M $\rightarrow$ NumInqLast6Mexcl7days \\
Net-Fraction-Revolving-Burden & NumRevolvingTradesWBalance $\rightarrow$ NetFractionRevolvingBurden \\
NetFractionInstallBurden & NumInstallTradesWBalance $\rightarrow$ NetFractionInstallBurden \\
Num-Revolving-Trades-W-Balance & NumTotalTrades $\rightarrow$ NumRevolvingTradesWBalance \\
Num-Install-Trades-W-Balance & NumTotalTrades $\rightarrow$ NumInstallTradesWBalance \\
NumBank-2-Natl-Trades-W-High-Utilization & NumRevolvingTradesWBalance $\rightarrow$ NumBank2NatlTradesWHighUtilization \\
PercentTradesWBalance & NumRevolvingTradesWBalance, NumInstallTradesWBalance, NumTotalTrades $\rightarrow$ PercentTradesWBalance \\
RiskPerformance & ExternalRiskEstimate, PercentTradesNeverDelq, NumTrades90Ever2DerogPubRec, MaxDelqEver, NetFractionRevolvingBurden, NumInqLast6M $\rightarrow$ RiskPerformance \\
\bottomrule
\end{tabular}
\caption{Extracted Feature Dependencies for \texttt{HELOC} Dataset}
\label{tab:dep_heloc}
\end{table*}

\begin{table*}[htbp]
\centering
\footnotesize
\renewcommand{\arraystretch}{1.1}
\begin{tabular}{p{0.28\linewidth} p{0.68\linewidth}}
\toprule
\textbf{Feature} & \textbf{Dependencies} \\
\midrule
Diabetes\_012 & HighBP, HighChol, BMI, Smoker, Stroke, HeartDiseaseorAttack, PhysActivity, Fruits, Veggies, HvyAlcoholConsump, GenHlth, MentHlth, PhysHlth, DiffWalk, Age, Sex, Income, Education $\rightarrow$ Diabetes\_012 \\
HighBP & BMI, Age, Sex, PhysActivity, HeartDiseaseorAttack, GenHlth, Income $\rightarrow$ HighBP \\
HighChol & BMI, Age, HighBP, PhysActivity, GenHlth, Income $\rightarrow$ HighChol \\
CholCheck & AnyHealthcare, Income, Education $\rightarrow$ CholCheck \\
BMI & PhysActivity, Fruits, Veggies, Sex, Age $\rightarrow$ BMI \\
Smoker & Age, Sex, Education, Income $\rightarrow$ Smoker \\
Stroke & HighBP, HeartDiseaseorAttack, Age, GenHlth, BMI $\rightarrow$ Stroke \\
HeartDiseaseorAttack & HighBP, HighChol, Stroke, Age, BMI $\rightarrow$ HeartDiseaseorAttack \\
PhysActivity & Age, Sex, Income, Education $\rightarrow$ PhysActivity \\
Fruits & Income, Education, Sex $\rightarrow$ Fruits \\
Veggies & Income, Education, Sex $\rightarrow$ Veggies \\
HvyAlcoholConsump & Sex, Age, Income $\rightarrow$ HvyAlcoholConsump \\
AnyHealthcare & Income, Education $\rightarrow$ AnyHealthcare \\
NoDocbcCost & Income, AnyHealthcare $\rightarrow$ NoDocbcCost \\
GenHlth & PhysHlth, MentHlth, DiffWalk $\rightarrow$ GenHlth \\
MentHlth & Income, Age, Sex $\rightarrow$ MentHlth \\
PhysHlth & Age, Income, Sex $\rightarrow$ PhysHlth \\
DiffWalk & Age, BMI, HeartDiseaseorAttack $\rightarrow$ DiffWalk \\
Sex & --- \\
Age & --- \\
Education & --- \\
Income & --- \\
\bottomrule
\end{tabular}
\caption{Extracted Feature Dependencies for \texttt{CDC Diabetes} Dataset.}
\label{tab:dep_diabetes}
\vspace{-10pt}
\end{table*}

\end{document}